\documentclass[lettersize,journal]{IEEEtran}
\usepackage[pdftex]{graphicx}
\DeclareGraphicsExtensions{.pdf,.jpeg,.png,.eps}
\usepackage{float}
\usepackage{subfigure}
\usepackage{bm}
\usepackage{amsmath,amsfonts}
\usepackage{multirow}
\usepackage{algorithmic}
\newcommand\algorithmicprocedure{\textbf{procedure}}
\newcommand{\algorithmicendprocedure}{\algorithmicend\ \algorithmicprocedure}
\makeatletter
\newcommand\PROCEDURE[3][default]{%
  \ALC@it
  \algorithmicprocedure\ {#2}(#3)%
  \ALC@com{#1}%
  \begin{ALC@prc}%
}
\newcommand\ENDPROCEDURE{%
  \end{ALC@prc}%
  \ifthenelse{\boolean{ALC@noend}}{}{%
    \ALC@it\algorithmicendprocedure
  }%
}
\newenvironment{ALC@prc}{\begin{ALC@g}}{\end{ALC@g}}
\makeatother
\usepackage{array}
\usepackage{textcomp}
\usepackage{stfloats}
\usepackage{url}
\usepackage{verbatim}
\usepackage{booktabs}
\usepackage{gensymb}
\usepackage{enumerate}
\usepackage{threeparttable}
\usepackage{pifont}
\usepackage{algorithm}
\usepackage{textcomp}

\def\BibTeX{{\rm B\kern-.05em{\sc i\kern-.025em b}\kern-.08em
    T\kern-.1667em\lower.7ex\hbox{E}\kern-.125emX}}
\usepackage{balance}
\usepackage{xcolor}
  
\newcommand{\rome}[1]{\uppercase\expandafter{\romannumeral#1}}
\newcommand{\etal}{\textit{et al.}}

\newcommand{\name}{ThinTact}
\begin{document}
\title{\name: Thin Vision-Based Tactile Sensor by Lensless Imaging}
\author{Jing Xu, \textit{Member, IEEE}, Weihang Chen, Hongyu Qian, Dan Wu, \textit{Member, IEEE}, Rui Chen, \textit{Member, IEEE}

\thanks{
\textit{(Jing Xu and Weihang Chen contributed equally to this work.) (Corresponding author: Rui Chen.)}.

Jing Xu, Weihang Chen, Hongyu Qian, Dan Wu, Rui Chen are with the Department of Mechanical Engineering, Tsinghua University, Beijing 100084, China (e-mail:
jingxu@tsinghua.edu.cn; chenwh14.thu@gmail.com; qhy23@mails.tsinghua.edu.cn; wud@tsinghua.edu.cn; chenruithu@tsinghua.edu.cn).
}}

\maketitle

\begin{abstract}
Vision-based tactile sensors have drawn increasing interest in the robotics community. However, traditional lens-based designs impose minimum thickness constraints on these sensors, limiting their applicability in space-restricted settings. In this paper, we propose \name, a novel lensless vision-based tactile sensor with a sensing field of over 200 mm${}^2$ and a thickness of less than 10 mm. \name ~utilizes the mask-based lensless imaging technique to map the contact information to CMOS signals.
To ensure real-time tactile sensing, we propose a real-time lensless reconstruction algorithm that leverages a frequency-spatial-domain joint filter based on discrete cosine transform (DCT). This algorithm achieves computation significantly faster than existing optimization-based methods. 
Additionally, to improve the sensing quality, we develop a mask optimization method based on the generic algorithm and the corresponding system matrix calibration algorithm.
We evaluate the performance of our proposed lensless reconstruction and tactile sensing through qualitative and quantitative experiments. Furthermore, we demonstrate \name's practical applicability in diverse applications, including texture recognition and contact-rich object manipulation.

\end{abstract}

\begin{IEEEkeywords}
Tactile sensing, lensless imaging, mask optimization, real-time reconstruction
\end{IEEEkeywords}

\section{Introduction}

\IEEEPARstart{T}{actile} sensing plays a crucial role in robot manipulation by providing necessary contact information~\cite{hogan2020tactile, she2021cable,liu2023enhancing,bian2023transtouch}. Vision-based tactile sensors transform the contact properties into images and have the advantages of high spatial resolution, low fabrication cost, and compatibility with learning-based computer vision techniques~\cite{yamaguchi2019recent, zhang2022hardware}. Utilizing vision-based tactile sensors, successful examples of robot manipulation are emerging~\cite{hogan2020tactile, she2021cable, wang2020swingbot, dong2021tactile}. 

However, in most unstructured environments where tactile sensing is necessary, objects or devices are typically designed for human use. As a result, they are well suited to the dimensions of human fingertips, which generally range from 15 to 20 mm in diameter. This poses significant challenges for vision-based tactile sensors because most of them use lens systems for image capturing, whose thickness is constrained by the minimum focusing distance~\cite{yuan2017gelsight, ward2018tactip, zhang2022tac3d}. This constraint significantly limits the potential applications of tactile sensors, such as parallel gripper grasping in cluttered scenarios, manipulating daily objects and integration with multi-finger hands.

\begin{figure}[t]
    \centering
    \includegraphics[width=8.5cm]{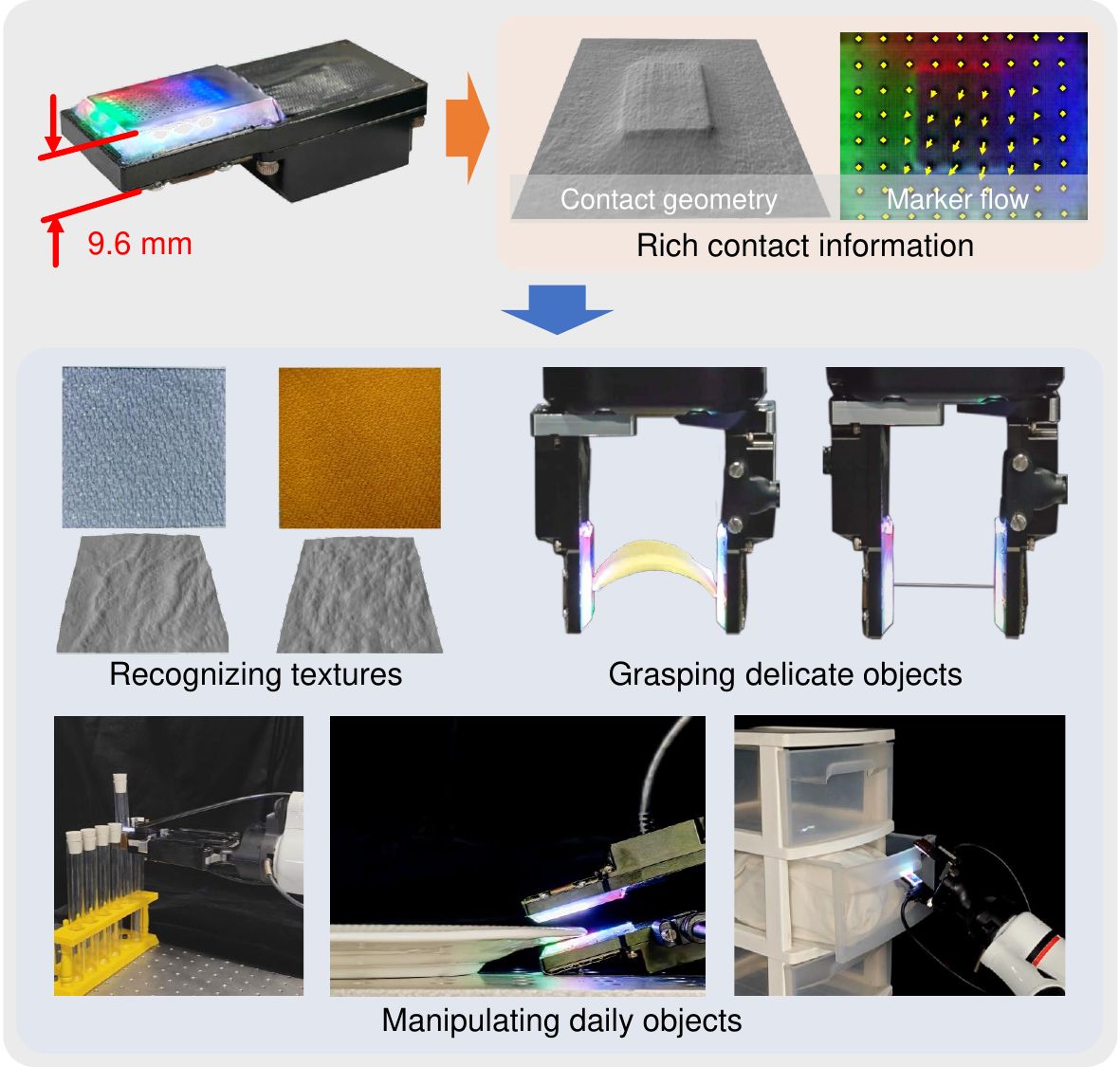}

    \caption{In this work, we propose \name, a thin vision-based tactile sensor with a thickness of less than 10 mm. To overcome the thickness constraint of the lens system, we utilize the amplitude-mask-based lensless imaging technique to translate the contact information into CMOS signals. We first reconstruct a clear image from the CMOS signal, and then compute the contact geometry and marker displacements. The high sensitivity and thin profile of ThinTact enables many applications, including texture recognition, delicate object grasping and object manipulation. }
    \label{fig:teaser}
\end{figure}

In this work, we aim to overcome the thickness constraint by eliminating the dependence of lens system. We propose \name, a novel vision-based tactile sensor with a thickness of 9.6 mm, which utilizes the amplitude-mask-based lensless imaging technique to map the contact information to CMOS signals. 
Fig.~\ref{fig:teaser} shows the sensor and its capabilities. 
Due to our use of the lensless imaging technique, the CMOS signal does not present a clear RGB image. For utilization, we first reconstruct the RGB image from the CMOS signal and then compute the contact geometry and marker displacements. Our sensor has high sensitivity, enabling it to reconstruct the detailed geometries of different textures and detect small contact forces during object grasping. Its thin profile also enables contact-rich object manipulation in confined spaces.

Incorporating lensless imaging into tactile sensors presents unique challenges. Firstly, lensless imaging requires a computational algorithm to reconstruct a clear image of the scene. The need for computational efficiency in closed-loop and reactive control of robot manipulation makes existing optimization-based approaches inapplicable. To address this, we propose a real-time lensless reconstruction algorithm. This algorithm employs a frequency-spatial-domain joint filter based on the Discrete Cosine Transform (DCT) to separate components in captured measurement images and reconstructs the scene using an analytical solution. The entire computation time is less than 2 ms, significantly faster than optimization-based methods~\cite{adams2017single}.
Secondly, constructing a thin tactile sensor with a large sensing field necessitates that the distance between the scene and the CMOS image sensor is smaller than the CMOS size, which is rare in most lensless applications. This makes traditional Maximum-Length-Sequence-based (MLS-based) masks~\cite{asif2016flatcam} unsuitable for the proposed sensor. To overcome this, we propose a mask optimization method based on fast lensless imaging simulation and the genetic algorithm (GA). The optimized mask yields reconstructed images with improved uniformity and quality compared to those based on MLS. Additionally, we propose a corresponding calibration algorithm for the optimized mask. Lastly, we employ the Real2Sim method~\cite{chen2022bidirectional} on \name\ to improve \name’s surface deformation measurement accuracy.

To validate the effectiveness of \name, we first characterize the resolution of \name's tactile image. Secondly, we evaluate the surface deformation measurement accuracy quantitatively. Lastly, we demonstrate the practicability of \name~in many applications, including texture recognition, grasping of delicate objects and contact-rich object manipulation in confined spaces.

In summary, the contributions of this work are as follows:
\begin{enumerate}
\item We propose a novel lensless vision-based tactile sensor design, which overcomes the thickness constraint imposed by the lens system and achieves 9.6 mm thickness.
\item We propose a real-time lensless reconstruction algorithm, which is significantly faster than optimization-based counterparts.
\item We propose a mask optimization which improves the reconstruction quality. Correspondingly, we design a calibration algorithm to compute system matrices of the optimized mask.
\item We conduct qualitative and quantitative experiments to validate the effectiveness of our proposed method and demonstrate the usefulness of our tactile sensor in various robot tasks.
\end{enumerate}

The remainder of this article is structured as follows. Section \ref{sec:related_works} reviews the relevant literature on tactile sensors and lensless imaging. Section \ref{sec:methods} describes the design of the proposed tactile sensor and the methodology for image reconstruction and tactile sensing. Section \ref{sec:experiments_fabrication} presents the detailed fabrication process of the sensor. Section \ref{sec:experiments_lensless_imaging_subsystem} evaluates the performance of the lensless imaging subsystem, and Section~\ref{sec:experiment_tactile_sensor} demonstrates the capabilities of the tactile sensor. Section~\ref{sec:applicatoin} examines the proposed tactile sensor's practicability in various robot applications. Finally, Section~\ref{sec:sec_conclusion} concludes this work.

\section{Related Work}
\label{sec:related_works}

\subsection{Vision-Based Tactile Sensing}\label{sec:related_works_tactile_sensing}
Tactile sensing has been increasingly applied in robot manipulation, especially for contact-rich scenarios~\cite{billard2019trends}. According to Wang \etal's review~\cite{wang2019tactile}, tactile sensors' principle can be categorized into the following modes: resistive, capacitive, piezoelectric, frictional, and optical mechanisms. Vision-based tactile sensors, a type of sensor that operates on optical mechanisms, have recently attracted significant attention. Their advantages lie in their superior spatial resolution, robustness against electromagnetic interference, cost-effectiveness, and the ability to capture rich contact information including simultaneous measurement of normal and tangential forces~\cite{yamaguchi2019recent,zhang2022hardware}. The GelSight sensor represents a notable example of vision-based tactile sensors~\cite{li2014localization,yuan2017gelsight,ma2019dense,taylor2022gelslim}. It captures high-resolution color images of the deformed elastomer surface which is illuminated from various angles. These images provide rich information of the contact surface, including 3D geometries and deformation. Aside from GelSight, there are other vision-based tactile sensors which can also provide rich contact information~\cite{sato2009finger,ward2018tactip,sferrazza2019design,sui2021incipient}. However, the common limitation of these vision-based tactile sensors is their large thickness and volume, which restricts the application in cluttered scenes and integration with dexterous grippers.

\subsection{Compact Tactile Sensor Design}
To address the aforementioned issue, compact tactile sensors have been developed.
A common approach involves the use of mirrors to fold the optical path, as seen in systems like GelSlim~\cite{donlon2018gelslim} and GelSight Wedge~\cite{wang2021gelsight}. However, despite that the camera is relocated, constraints in sensor integration persist due to the considerable scene-to-CMOS distance. Several studies have comprehensively designed robotic fingers or grippers with tactile sensing~\cite{padmanabha2020omnitact,romero2020soft,gomes2020geltip,sun2022soft,azulay2023allsight}. These designs allocate substantial space for optical transmission. Consequently, altering their form factor for integration into other robotic components poses a significant challenge. Other works have developed sensors based on small-sized cameras~\cite{lambeta2020digit} or wide-angle cameras~\cite{taylor2022gelslim} and achieve impressive results. However, the resulting thickness still exceeds that of human fingers. In summary, none of these sensors have a thickness of less than 20 mm, making it challenging to achieve human-level dexterity.

Recently, several vision-based tactile sensors have adopted multi-lens imaging systems instead of the traditional single-lens models~\cite{shimonomura2013combined,yu2023omsense,song2021bionic,zhang2022multidimensional,chen2022thin}. Despite the flexibility offered by multi-lens systems, most of them still possess a bulky size. Therefore, our focus is on the work of Chen \etal~\cite{chen2022thin}, who proposed a tactile sensor with only 5 mm thickness. The sensor utilizes microfabricated micro-lens arrays (MLA) for imaging. However, each micro-lens still adheres to the constraints of object distance, image distance, and magnification ratio, leading to overlapping sub-image areas and a low utilization rate of the image sensor. Additionally, the process of fabricating the MLA and preventing crosstalk introduces higher manufacturing costs.

In this work, we propose a novel, compact, vision-based tactile sensor that utilizes lensless imaging principles. This approach eliminates the constraints in traditional imaging systems and results in a sensor with a thickness of less than 10 mm and a sensing field of over 200 mm$^2$. By employing a simple binary mask composed of a glass substrate and a thin chrome film, our sensor offers ease of manufacturing and cost-effectiveness. 

\subsection{Lensless Imaging}
\label{sec:related_works_lensless_imaging}
In a lensless camera, an optical encoder replaces the traditional lens to encode scene information into sensor measurements. The scene is later recovered using appropriate algorithms. We refer to \cite{boominathan2022recent} for a comprehensive review of lensless imaging. In the context of tactile sensing, the appealing advantages of lensless imaging include reduced camera size and large field of view (FOV), because lensless imaging breaks the constraint between working distance and FOV in traditional lens-based imaging.

Designing a lensless imaging system is a systematic task, as the imaging conditions, the choice and design of the optical encoder, the reconstruction algorithm, are closely coupled. To make lensless imaging applicable in tactile sensing, several issues must be addressed. Firstly, the computational burden should be minimized to ensure real-time measurement. Secondly, to design a tactile sensor with a large sensing area and low thickness, the scene-to-CMOS distance is smaller than the CMOS size, which is uncommon in traditional lensless imaging applications.

With these considerations in mind, we delve into the field of lensless imaging. One method to reduce the computational burden is to use a convolutional imaging model, such as DiffuserCam~\cite{antipa2018diffusercam} and PhlatCam~\cite{boominathan2020phlatcam}. This method requires the point spread function (PSF) to be spatially invariant across the entire scene. However, this assumption does not hold when the scene-to-CMOS distance is smaller than the CMOS size. Although \cite{kuo2020chip} proposed a model to account for this situation, it cannot be solved efficiently in real time. Another method to realize real-time reconstruction is to use a separable amplitude mask, known as FlatCam~\cite{asif2016flatcam}. Adams \etal~have extended FlatCam to close-up imaging situations~\cite{adams2017single}, but the corresponding imaging model cannot be solved analytically and efficiently. In this work, we adopt the separable amplitude mask and propose a novel filter to decompose the sensor measurements and obtain the coding component, enabling efficient real-time scene reconstruction. 
It is noteworthy that recent studies have suggested the application of deep neural networks for lensless image reconstruction~\cite{ monakhova2019learned, khan2020flatnet, zeng2021robust, bagadthey2022flatnet3d}. However, in the context of compact tactile sensors, it is difficult to gather scene-measurement-paired datasets for training. Furthermore, these learning-based methods often encounter hallucination artifacts that lack high-fidelity~\cite{khan2020flatnet}. Therefore, in this work, we have designed the proposed filter based on mathematical and physical properties. This approach not only ensures its generalizability but also guarantees a lower computational burden compared to deep neural networks.

\begin{figure*}[t]
    \centering
    \includegraphics[width=18cm]{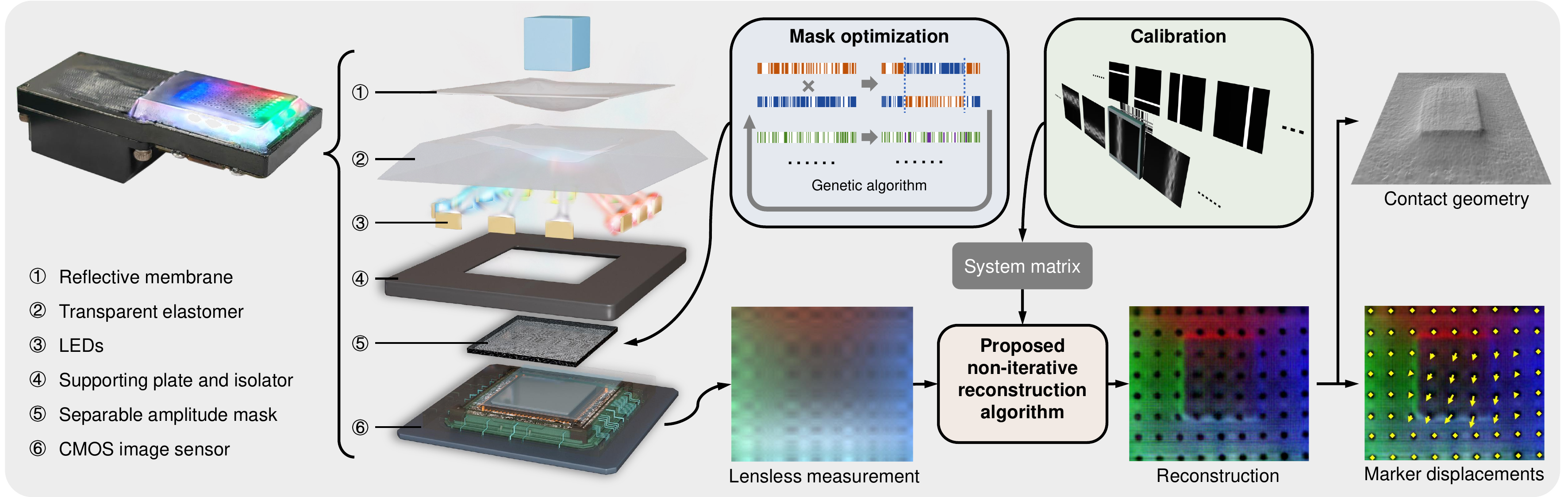}
    \caption{Workflow of \name. On the left of this figure is the schematic of \name. It utilizes a separable amplitude mask for imaging. To address the challenging problems associated with integrating lensless imaging into tactile sensing, we have developed a non-iterative real-time reconstruction algorithm, a mask optimization method, and a corresponding calibration algorithm. The reconstructed images allow us to determine contact geometry and marker displacements, which helps robot manipulation.}
    \label{fig:work_flow}
\end{figure*}

\section{Methodology}
\label{sec:methods}

To address the inherent thickness limitations of lens-based imaging systems in existing vision-based tactile sensors, we present a novel tactile sensor called \name. Our proposed approach leverages lensless imaging techniques to capture and map contact information into CMOS signals. In the left part of Fig.~\ref{fig:work_flow}, we illustrate the design of our tactile sensor.
It consists of a transparent elastomer covered by a reflective membrane, a separable amplitude mask, a CMOS sensor and LEDs of different colors.
Light emitted from the LEDs is reflected on the membrane, constructing a colored 2D ``scene'' $\bm{X} \in \mathbb{R}^{N\times M \times 3}$. Subsequently, the light transports through the mask and reaches the CMOS, forming a measured RGB image $\bm{Y} \in \mathbb{R}^{R\times S \times 3}$. In lensless imaging, the R, G and B channels can be handled separately, therefore in next discussions we neglect the channel dimension for simplicity, i.e., $\bm{X} \in \mathbb{R}^{N\times M}$, $\bm{Y} \in \mathbb{R}^{R\times S}$.

Fig.~\ref{fig:work_flow} illustrates the workflow of \name.
When an object comes into contact with the sensor, the sensing surface deforms, which leads to the change of the scene $\bm{X}$ and the measurement $\bm{Y}$. We first describe the lensless imaging model that transforms $\bm{X}$ into $\bm{Y}$ in Sec.~\ref{sec:imaging_model}. In order to reconstruct $\bm{X}$ from $\bm{Y}$ efficiently in real time, we propose a novel non-iterative reconstruction algorithm (Sec.~\ref{sec:proposed_reconstruction}), which is approximately 1000 times faster than the optimization-based reconstruction algorithm~\cite{adams2017single}. 
However, we observed that our proposed algorithm's reconstruction quality deteriorates for typical separable masks~\cite{asif2016flatcam}, when the distance between the scene and the CMOS is small. As our sensor operates under such conditions, we address this issue by presenting a mask pattern optimization algorithm (Sec.~\ref{sec:pattern optimization}).
The calibration process of the lensless system matrices is presented in Sec.~\ref{sec:calibration_method}.
Once the scene $\bm{X}$ is reconstructed, the surface deformation can be computed by photometric stereo and marker tracking (Sec.~\ref{sec:depth_recon}).

\subsection{Lensless Imaging Model}
\label{sec:imaging_model}

In this work, we choose to use a 2D separable amplitude mask as the coded aperture, which has the advantage of better computational efficiency and lower fabrication cost. 
The binary mask pattern $\bm{M} \in \{0,1\}^{K\times K}$ is constructed by:

\begin{equation}
    \bm{M} = \frac{\bm{1}\cdot\bm{1}^\top+\bm{\varphi}_1\bm{\varphi}_2^{\top}}{2}
    \label{eq:mask_generation}
\end{equation}
where $\bm{\varphi}_1,\bm{\varphi}_2$ are two vectors, each entry of which is either -1 or +1. In this work, we assume $\bm{\varphi}_1=\bm{\varphi}_2=\bm{\varphi}$. An example of the separable mask is shown in Fig.~\ref{fig:work_flow}.

\begin{figure}
    \centering
    \subfigure[]{
		\includegraphics[height=2.5cm]{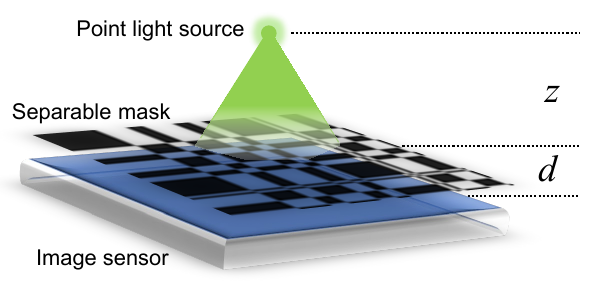}
		\label{fig:t2s_model_a}
	}
    \subfigure[]{
	\includegraphics[height=2.5cm]{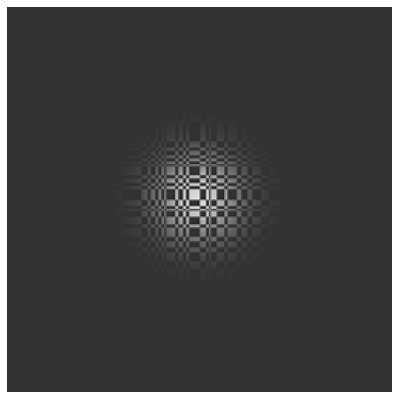}
	\label{fig:t2s_model_b}
	}
     \subfigure[]{
	\includegraphics[height=2.5cm]{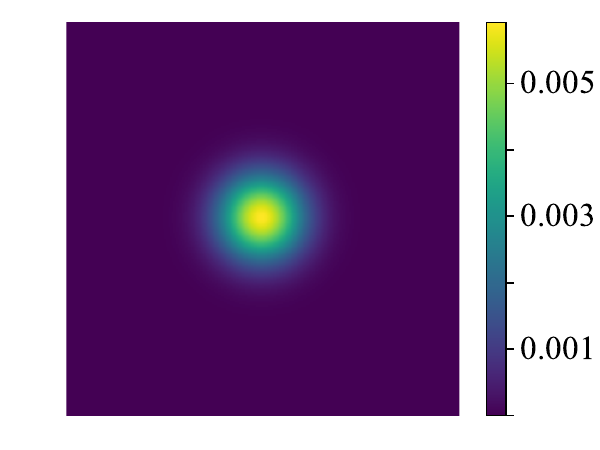}
	\label{fig:t2s_model_c}
	}
     \subfigure[]{
	\includegraphics[height=2.5cm]{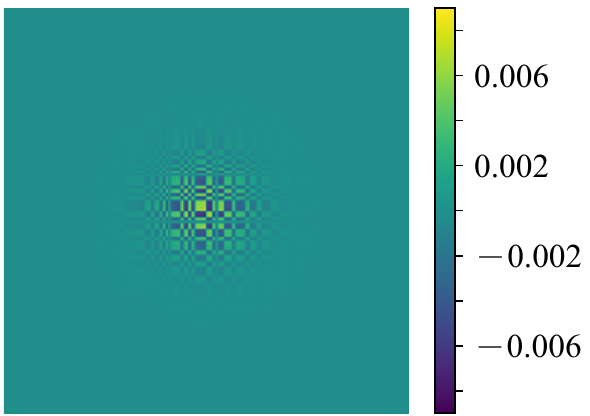}
	\label{fig:t2s_model_d}
	}
     \subfigure[]{
	\includegraphics[height=3cm]{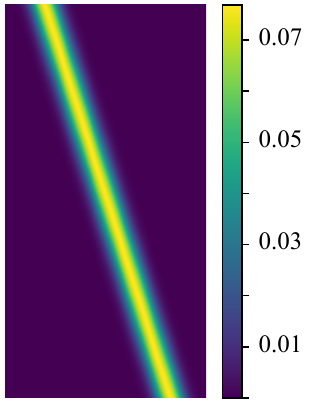}
	\label{fig:t2s_model_e}
	}
     \subfigure[]{
	\includegraphics[height=3cm]{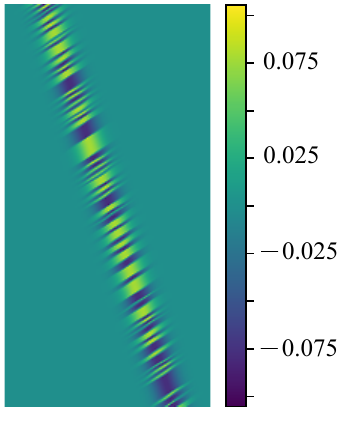}
	\label{fig:t2s_model_f}
	}
     \caption{In the proposed tactile sensor, the scene is close to the CMOS. (a) Close-up imaging situation. In this case, the scene-to-CMOS distance is smaller than the sensor size so that the PSF only covers a small area. (b) The resulting PSF. In the T2S model, the PSF can be decomposed into two terms: (c) the open component  and (d) the coding component. (e) System matrix $\bm{P}_\text{o}$. (f) System matrix $\bm{P}_\text{c}$.}
    \label{fig:t2s_model}
\end{figure}

Due to the limited thickness of our proposed tactile sensor, the distance between the scene and the CMOS sensor is small. Consider a point light source present in the scene (Fig. \ref{fig:t2s_model_a}). The light source casts a portion of the mask pattern onto the CMOS (Fig. \ref{fig:t2s_model_b}), which is referred to as the Point Spread Function (PSF).
The PSF can be decomposed into two components:
The first component (Fig. \ref{fig:t2s_model_c}) corresponds to the term $\bm{1}\cdot\bm{1}^\top/2$ in the mask. This component approximates the response of the CMOS when it is exposed to an ``open" mask, i.e., a mask without any apertures.
The second component (Fig. \ref{fig:t2s_model_d}) corresponds to the term $\bm{\varphi}\bm{\varphi}^{\top}/2$ in the mask. This component represents the coding effect of the mask and captures the features introduced by the mask pattern.

Because the measurement $\bm{Y}$ is a superposition of the PSFs of all the points in the scene 
$\bm{X}$, the imaging model can be expressed as
\begin{equation}
  \bm{Y} = \bm{P}_\text{o}\bm{X}\bm{Q}_\text{o}^{\top} + \bm{P}_\text{c}\bm{X}\bm{Q}_\text{c}^{\top}
  \label{eq:T2S_model}
\end{equation}
where $\bm{P}_\text{o}\in \mathbb{R}^{R\times N}$, $\bm{Q}_\text{o}\in \mathbb{R}^{S\times M}$, $\bm{P}_\text{c}\in \mathbb{R}^{R\times N}$ and $\bm{Q}_\text{c}\in \mathbb{R}^{S\times M}$ are the system matrices, and the subscripts ``o" and ``c" denote ``open" and ``coding", respectively. 
We present some typical system matrices in Fig. \ref{fig:t2s_model_e} and Fig. \ref{fig:t2s_model_f}. 
This model is known as the Texas Two-Step (T2S) model \cite{adams2017single}.

Given the imaging model, the scene $\bm{X}$ can be reconstructed by solving the following optimization problem:
\begin{equation}
  \hat{\bm{X}}=\text{argmin}_{\bm{X}}\Vert\bm{P}_\text{o}\bm{X}\bm{Q}_\text{o}^\top+\bm{P}_\text{c}\bm{X}\bm{Q}_\text{c}^\top-\bm{Y}\Vert_\text{F}^2+\tau\Vert\bm{X}\Vert_\text{F}^2
\label{eq:T2S_optimization}
\end{equation}
where $\tau$ is the regularization coefficient.
% Unfortunately, currently 
Unfortunately, there are no computationally feasible analytical solutions at present. Instead, it can be solved using iterative methods, such as Nesterov gradient method \cite{nesterov2012efficiency} as introduced in \cite{adams2017single}, whose time consumption typically ranges from a fraction of a second to a few seconds on GPU, which makes it inapplicable in tactile sensing. To address this issue, in the next subsection, we propose a non-iterative reconstruction algorithm to greatly decrease the computational time and enable real-time image reconstruction.

\subsection{Non-Iterative Real-Time Lensless Image Reconstruction}
\label{sec:proposed_reconstruction}
\begin{figure}
    \centering
    \includegraphics[width=8.5cm]{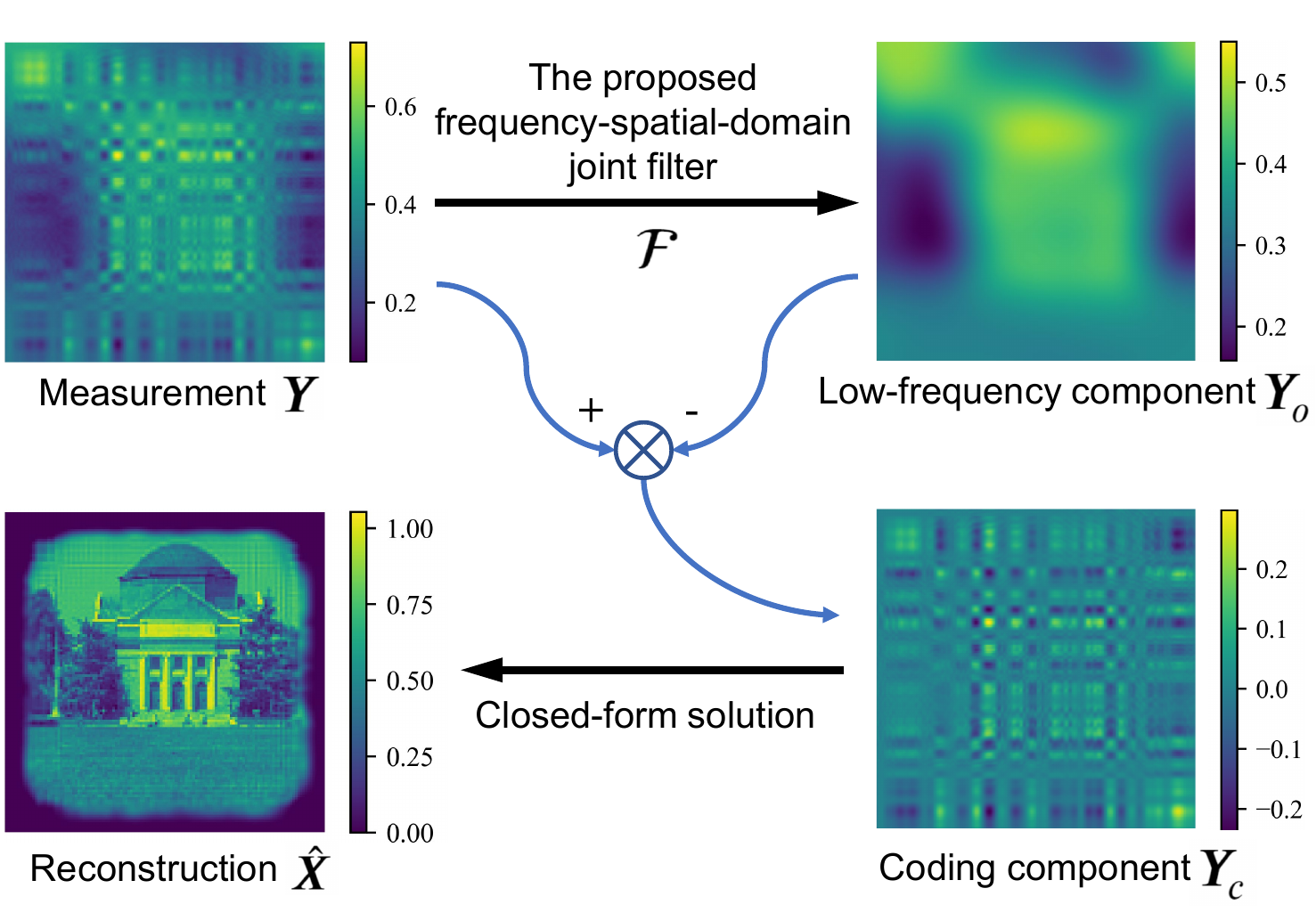}
    \caption{The pipeline of the proposed non-iterative reconstruction algorithm.}
    \label{fig:reconstruction_pipeline}
\end{figure}

In this section, we propose a non-iterative real-time reconstruction algorithm. It first utilizes a frequency-spatial-domain joint filter to remove the $\bm{P}_\text{o}\bm{X}\bm{Q}_\text{o}^{\top}$ term.
The remaining term is then solved analytically. Fig. \ref{fig:reconstruction_pipeline} shows the proposed algorithm.

\subsubsection{Frequency-spatial-domain joint filter}
In contrast to the T2S model, the imaging model  $\bm{Y} = \bm{P}_\text{c}\bm{X}\bm{Q}_\text{c}^{\top}$ in ~\cite{asif2016flatcam} can be solved analytically. However, the reduced model is only applicable to situations with large scene-to-CMOS distances and cannot be used for scene reconstruction in our tactile sensor.
The key of the proposed method is a filter $\mathcal{F}$ which obtains the low-frequency component $\bm{P}_\text{o}\bm{X}\bm{Q}_\text{o}^{\top}$ from the measurement $\bm{Y}$ so that the coding component $\bm{P}_\text{c}\bm{X}\bm{Q}_\text{c}^{\top}$ can be obtained by subtraction:

\begin{align}
  \bm{Y}_\text{o} &= \mathcal{F}(\bm{Y})\\
  \bm{Y}_\text{c} &= \bm{Y}-\bm{Y}_\text{o} =  \bm{Y}-\mathcal{F}(\bm{Y})
  \label{eq:low_comp_filter}
\end{align}
where $\bm{Y}_\text{o}$ denotes $\bm{P}_\text{o}\bm{X}\bm{Q}_\text{o}^{\top}$ and $\bm{Y}_\text{c}$ denotes $\bm{P}_\text{c}\bm{X}\bm{Q}_\text{c}^{\top}$. From the mapping $\mathcal{F}$'s effect we know that it is linear, and therefore we should design a linear filter to ensure generalizability.

To design $\mathcal{F}$, we look into its physical characteristics. As previously discussed in Sec. \ref{sec:imaging_model}, the term $\bm{Y}_\text{o}$ approximates the response in the absence of a mask, while the term $\bm{Y}_\text{c}$ reflects the modulation introduced by the mask. Consequently, $\bm{Y}_\text{o}$ primarily contains low-frequency components. On the other hand, although $\bm{Y}_\text{c}$ mainly contains high-frequency components, it also includes low-frequency components. These can intermingle with $\bm{Y}_\text{o}$, with the mixture ratio varying spatially and determined by the mask. Therefore, the design of the filter should take into account both the frequency and spatial domains.
    
Based on the above discussion, we design the following two basic units to form a frequency-spatial-domain joint filter. The frequency-domain filter unit is:
\begin{equation}
  \mathcal{F}_f(\bm{Y}) = \text{DCT}^{-1}[\bm{\Phi}_f \odot \text{DCT}(\bm{Y})]
\end{equation}
The spatial-domain filter unit is:
\begin{equation}
  \mathcal{F}_s(\bm{Y}) = \bm{\Phi}_s \odot \bm{Y}
\end{equation}
where $\bm{\Phi}_f\in \mathbb{R}^{R\times S}$ is the parameter matrix of the frequency-domain filter, $\text{DCT}(\bullet)$ and $\text{DCT}^{-1}(\bullet)$ stand for Discrete Cosine Transform (DCT) and its inverse transform, $\odot$ is element-wise product; $\bm{\Phi}_s\in \mathbb{R}^{R\times S}$ is the parameter matrix of the spatial-domain filter. DCT is widely adopted in image and video compression, and is known for its excellent energy compaction~\cite{khayam2003discrete}. We also empirically find that DCT performs better than Discrete Fourier Transform (DFT) in this case.

By cascading them together, we obtain the proposed frequency-spatial-domain joint linear filter:
\begin{equation}
  \mathcal{F}(\bm{Y}) = \mathcal{F}_s(\mathcal{F}_f(\bm{Y}))=\bm{\Phi}_s \odot\text{DCT}^{-1}[\bm{\Phi}_f \odot \text{DCT}(\bm{Y})]
\end{equation}

Theoretically, when the arrangement of a separable-mask-based lensless imaging system is determined, the four system matrices $\bm{P}_\text{o}$, $\bm{Q}_\text{o}$, $\bm{P}_\text{c}$ and $\bm{Q}_\text{c}$ are also determined correspondingly and can be obtained through calibration, thereby determining the filter parameters $\bm{\Phi}_f$ and $\bm{\Phi}_s$. However, in practice, it is difficult to obtain the filter parameters through the system matrices analytically. Therefore, this paper uses the system matrices and multiple scene images to generate a virtual dataset, and performs gradient descent training on the proposed filter to obtain optimal $\bm{\Phi}_f$ and $\bm{\Phi}_s$.

\subsubsection{The closed-form solution of the scene}
Using the previously introduced filter, we obtain the coding component $\bm{Y}_\text{c}=\bm{P}_\text{c}\bm{X}\bm{Q}_\text{c}^{\top}=\bm{Y}-\mathcal{F}(\bm{Y})$. Then, the reconstruction of the scene can be calculated using a closed-form solution~\cite{asif2016flatcam}. Here we briefly provide the solution. For detailed derivation, please refer to \cite{asif2016flatcam}.

Reconstruction of the scene is formulated as the following optimization problem:
\begin{equation}
  \hat{\bm{X}}=\text{argmin}_{\bm{X}}\Vert\bm{P}_\text{c}\bm{X}\bm{Q}_\text{c}^\top-\bm{Y}_\text{c}\Vert_\text{F}^2+\tau\Vert\bm{X}\Vert_\text{F}^2
  \label{eq:optimization_tik}
\end{equation}
which can be solved by differentiating it and letting the gradient be zero:
\begin{equation}  \bm{P}_\text{c}^\top\bm{P}_\text{c}\bm{X}\bm{Q}_\text{c}^\top\bm{Q}_\text{c}+\tau\bm{X}=\bm{P}_\text{c}^\top\bm{Y}_\text{c}\bm{Q}_\text{c}
  \label{eq:gradient_equals_zero}
\end{equation}
To solve \eqref{eq:gradient_equals_zero}, first calculate the Singular Value Decomposition (SVD) of $\bm{P}_\text{c}$, $\bm{Q}_\text{c}$: $\bm{P}_\text{c}=\bm{U}_P\bm{\Sigma}_P\bm{V}_P^\top$, $\bm{Q}_\text{c}=\bm{U}_Q\bm{\Sigma}_Q\bm{V}_Q^\top$, and use $\bm{\sigma}_P$ and $\bm{\sigma}_Q$ to denote the vectors formed by the diagonal elements of $\bm{\Sigma}_P^2$ and $\bm{\Sigma}_Q^2$. The final solution can be written as
\begin{equation}
  \hat{\bm{X}} =\bm{V}_P [(\bm{\Sigma}_P \bm{U}_P^\top \bm{Y}_\text{c}\bm{U}_Q\bm{\Sigma}_Q ).\big/(\bm{\sigma}_P\bm{\sigma}_Q^\top + \tau\bm{1}\bm{1}^\top)]\bm{V}_Q^\top
  \label{eq:analytical_solution}
\end{equation}
where $.\big/$ is element-wise division of matrices. Note that only $\bm{Y}_\text{c}$ is obtained from $\bm{Y}$ from the image sensor in real time, while all other matrices can be calculated in advance. Therefore, fast reconstruction is possible using this closed-form solution.

\subsection{Mask Pattern Optimization}
\label{sec:pattern optimization}
In existing separable-mask-based lensless imaging systems, masks are typically designed based on vectors with specific mathematical properties, such as MLS~\cite{asif2016flatcam}. The advantage of MLS is that, the transfer matrix of the mask it forms will have large and slowly-decaying singular values, which is beneficial for image reconstruction. However, when the scene is close to the CMOS sensor, the system matrices begin to exhibit difference properties because the PSF only covers a small area. This change can result in reconstructed scenes suffering from nonuniform intensity or low Signal-to-Noise Ratio (SNR). To this end, we propose to generate the optimal mask pattern for such scenario using the genetic algorithm. Since the time consumption of the genetic algorithm largely depends on the time of each generation, we develop a method that directly generates corresponding system matrices from the vector, which significantly accelerates the optimization process.

\subsubsection{Mask optimization using genetic algorithm}
Given the assumption of mask symmetry in our work, the objective of mask optimization is to search for the optimal vector $\bm{\varphi}$.
Since each element of the vector $\bm{\varphi}\in \{-1, 1\}^K$ is binary, we choose the genetic algorithm~\cite{mitchell1998introduction} to solve the mask optimization problem. Here we present some critical design choices:

\noindent \textbf{Definition of genes.} The vector $\bm{\varphi}$ naturally becomes the gene. During optimization, the length of $\bm{\varphi}$ is fixed to be $K$ and each of its element can be 1 or -1.

\noindent \textbf{The crossover operator.} When performing crossover, we randomly select two corresponding loci on two gene sequences, and then exchange the gene fragments between the two loci. 

\noindent \textbf{The mutation operator.} When performing mutation, we randomly select $s$ loci on the gene sequence, and then invert them. The value of $s$ is determined based on the length of the gene. 

\noindent \textbf{The fitness function.} We use the quality of the simulated reconstructed image as the fitness function. For a certain ground truth scene image $\bm{X}_\text{GT}$, because our proposed reconstruction algorithm does not rely on the low-frequency component $\bm{P}_\text{o}\bm{X}_\text{GT}\bm{Q}_\text{o}^\top$, its simulated measurement image $\bm{Y}_{\text{c,sim}}$ under the mask corresponding to the gene $\bm{\varphi}$ can be calculated as:
      \begin{equation}
    \bm{Y}_{\text{c,sim}}(\bm{\varphi},\bm{X}_\text{GT})=\bm{P}_\text{c}(\bm{\varphi})\bm{X}_\text{GT}\bm{Q}_\text{c}(\bm{\varphi})^\top + \bm{\Sigma}
  \end{equation}
where $\bm{P}_\text{c}(\bm{\varphi})$ and $\bm{Q}_\text{c}(\bm{\varphi})$ are the system matrices generated from the gene $\bm{\varphi}$ and $\bm{\Sigma}$ is the pixel-wise noise. The method for directly generating the system matrices will be described in the following subsection. The next step is to use \eqref{eq:analytical_solution} to reconstruct the scene from $\bm{Y}_{\text{c,sim}}(\bm{\varphi},\bm{X}_\text{GT})$ and obtain the simulated reconstructed image $\hat{\bm{X}}_\text{sim}(\bm{\varphi},\bm{X}_\text{GT})$. Because a regularization coefficient is introduced when calculating the reconstructed image, the noise resistance of the system matrix can be judged by the quality of the reconstructed image. Two commonly used indicators in image quantitative evaluation indicators - Peak Signal-to-Noise Ratio (PSNR) and Structural Similarity Index Measure (SSIM) - are used to evaluate the quality of the reconstructed image,
  \begin{align}
    f_\text{SSIM}(\bm{\varphi},\bm{X}_\text{GT})&=\frac{1}{1-\text{SSIM}(\hat{\bm{X}}_\text{sim}(\bm{\varphi},\bm{X}_\text{GT}),\bm{X}_\text{GT})} \\
    f_\text{PSNR}(\bm{\varphi},\bm{X}_\text{GT})&=\text{PSNR}(\hat{\bm{X}}_\text{sim}(\bm{\varphi},\bm{X}_\text{GT}),\bm{X}_\text{GT})
  \end{align}
In order to avoid the nonuniform intensity phenomenon in the reconstructed image, an additional gradient-based evaluation indicator is added for the reconstruction of a pure white image $\bm{X}_\text{white}=\bm{1}^{N\times M}$:
  \begin{equation}
  \begin{aligned}
          f_\text{GRAD}(\bm{\varphi},\bm{X}_\text{white})=&2MN\big/\{\Vert \bm{G}_u(\hat{\bm{X}}_\text{sim}(\bm{\varphi},\bm{X}_\text{white}))\Vert_1 + \\ &\Vert \bm{G}_v(\hat{\bm{X}}_\text{sim}(\bm{\varphi},\bm{X}_\text{white}))\Vert_1\}
  \end{aligned}
  \end{equation}
where $ \bm{G}_u(\bullet)$ and $\bm{G}_v(\bullet)$ represent the gradient matrix in both directions.
The selection of specific parameters will be introduced in the experiment section.

\subsubsection{Direct generation of system matrices}
\begin{figure}
    \centering
    \subfigure[]{
    \includegraphics[width=2.8cm]{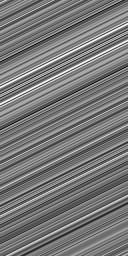}
    \label{fig:sysmat_analysis_a}
    }
    \hfill
    \subfigure[]{
    \includegraphics[width=5.2cm]{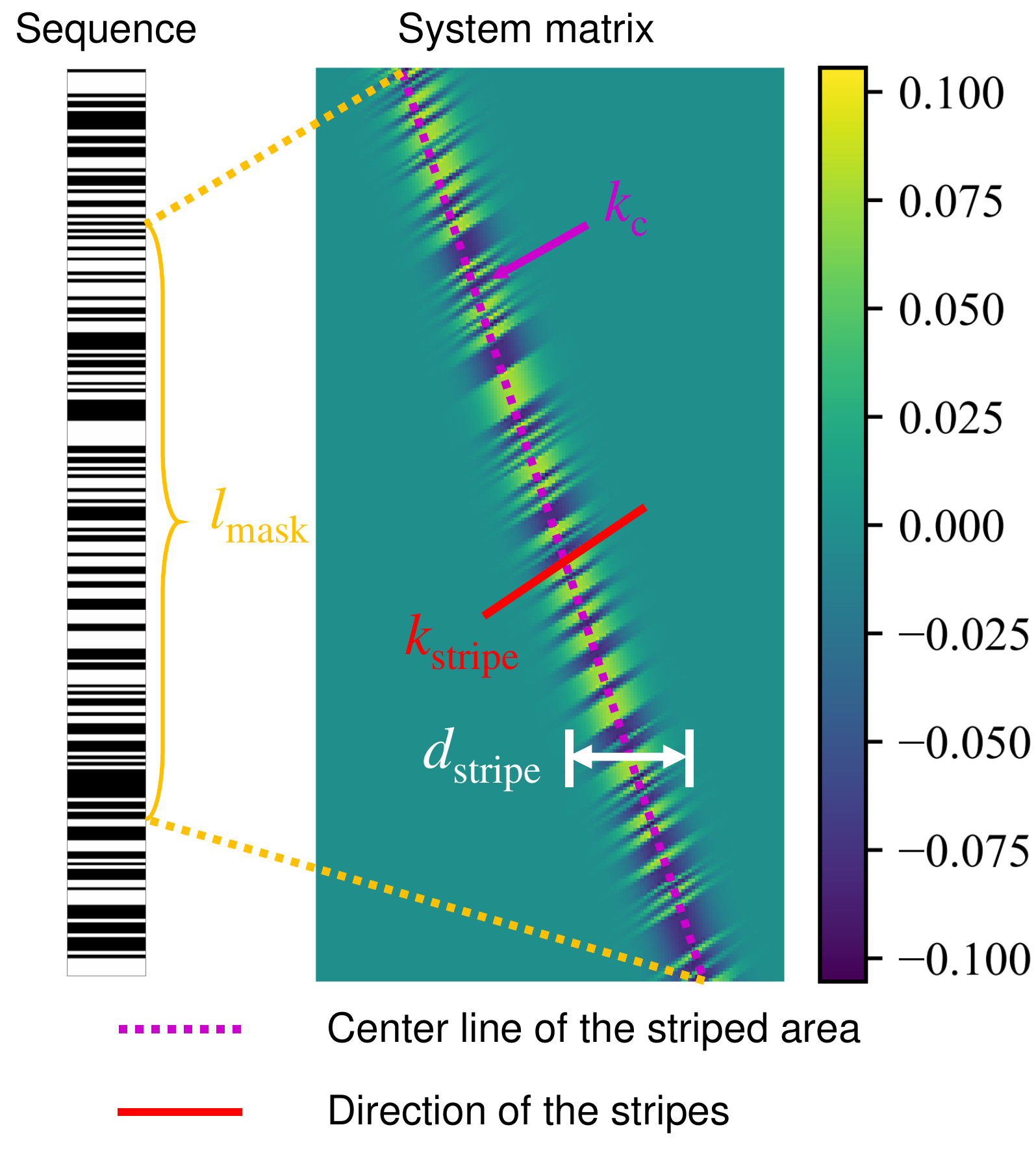}
    \label{fig:sysmat_analysis_b}
    }
    \caption{Analysis of system matrices. (a) Randomized system matrix generated in \cite{khan2020flatnet}. In this work, we generate system matrices that are suitable for close-up imaging scenarios. (b) Analysis of the system matrix. The vector on the left is used to form the mask and the corresponding system matrix is on the right. They are assumed to be aligned at their centers. The matrix has several key parameters: the slope $k_\text{c}$ of the center line of the striped area (the purple dashed line), the slope $k_\text{stripe}$ of the direction of the stripes (the red solid line), the width of the striped area $d_\text{stripe}$, and the effective length $l_\text{mask}$ in the vector (marked by the yellow dashed lines).}
    \label{fig:sysmat_analysis}
\end{figure}

\cite{khan2020flatnet} introduced a method to generate randomized system matrices for network training, as shown in Fig. \ref{fig:sysmat_analysis_a}. However, it cannot accommodate the imaging scenarios in our tactile sensor, and the values in the generated system matrix are randomized and thus cannot be used for simulation. In this work, we first extend the method to accommodate close-up imaging situations. Second, we ensure that the values in the generated system matrices are physically reasonable, enabling us to directly simulate the reconstructed scenes.

Without loss of generalizability, we take $\bm{P}_\text{c}$ as an example to explain the properties of the system matrices in separable-mask-based lensless imaging. As shown in Fig. \ref{fig:sysmat_analysis_b}, a vector to form the separable mask and the corresponding system matrix are shown. An important characteristic of the system matrix is the gradual fading of the striped area at its boundary, rather than covering the entire matrix. This unique aspect distinguishes the close-up imaging scenario. The fading effect is attributed to the limited response of each CMOS pixel when the light source is close to the CMOS. To approximate this fading effect in practice, we employ a sigmoid function. The width of the striped area, denoted as $d_\text{stripe}$, is experimentally determined to align with the actual system matrix. For other important parameters, as shown in Fig. \ref{fig:sysmat_analysis_b}, let $k_\text{c}$ denote the slope of the center line of the stripe area, $k_\text{stripe}$ denote the slope of the direction of stripes and $l_\text{mask}$ denote the effective physical length of the mask, corresponding to the center line in the system matrix. These values can be calculated as:

\begin{align}
  k_{\text{c}} &= \frac{\delta_{\text{scene}}}{\delta_{\text{sensor}}}  \label{eq:generate_sysmat_1}\\
  k_{\text{stripe}} &= -\frac{d\delta_{\text{scene}}}{z\delta_{\text{sensor}}} \label{eq:generate_sysmat_2} \\
  l_{\text{mask}} &= N\delta_{\text{sensor}} \label{eq:generate_sysmat_3}
\end{align}
where $\delta_{\text{sensor}}$ is the pixel spacing of the image sensor, $\delta_{\text{scene}}$ is the pixel size of the reconstructed scene, $d$ is the mask-to-CMOS distance, $z$ is the scene-to-mask distance, $N$ is the height of the measurement image for $\bm{P}_\text{c}$. After obtaining the above relations, we can use affine transformation to generate system matrices based on the vectors that form the mask pattern.

\subsection{Calibration of Lensless System Matrices}\label{sec:calibration_method}

The objective of calibration is to obtain the system matrices $\bm{P}_\text{o}$, $\bm{Q}_\text{o}$, $\bm{P}_\text{c}$ and $\bm{Q}_\text{c}$.
In \cite{adams2017single}, since the MLS vector satisfies $\bm{1}^\top\bm{\varphi}=0$, the corresponding columns of $\bm{P}_\text{o}$ and $\bm{P}_\text{c}$ are orthogonal. Thus, an SVD-based calibration method is adopted.

In our case, because our mask is optimized through the genetic algorithm and does not have such constraint, the calibration algorithm in \cite{adams2017single} is not applicable. This section presents our calibration algorithm.
\subsubsection{Collection of calibration images} 
The calibration images are the measurement images of a scanning horizontal slit and a scanning vertical slit, which are the same as \cite{adams2017single}. Take the measurement images of a scanning horizontal slit as an example. A horizontal slit is a scene of which only the $i$th row is activated, i.e., $\bm{X}_{\text{h}i}=\bm{e}_i\bm{1}^\top$, where $\bm{e}_i$ is a column vector of zeros with only the $i$th element to be 1. The corresponding measurement image $\bm{Y}_{\text{h}i}$ can be written as
\begin{equation}
\begin{aligned}
        \bm{Y}_{\text{h}i}&=(\bm{P}_\text{o}\bm{e}_i)(\bm{Q}_\text{o}\bm{1})^\top + (\bm{P}_\text{c}\bm{e}_i)(\bm{Q}_\text{c}\bm{1})^\top\\&=\bm{p}_{\text{o}i}\bm{q}_{\text{o}}^\top+\bm{p}_{\text{c}i}\bm{q}_{\text{c}}^\top
\end{aligned}
\end{equation}
where $\bm{p}_{\text{o}i}$ and $\bm{p}_{\text{c}i}$ are the $i$th columns of $\bm{P}_\text{o}$ and $\bm{P}_\text{c}$, respectively, and $\bm{q}_{\text{o}}$ and $\bm{q}_{\text{c}}$ are the sums of rows of $\bm{Q}_\text{o}$ and $\bm{Q}_\text{c}$, respectively. Note that $\bm{q}_{\text{o}}$ and $\bm{q}_{\text{c}}$ remain constant when $i$ changes. This property is important for the calibration algorithm. For a scanning vertical slit, we also obtain
\begin{equation}
\bm{Y}_{\text{v}i}=\bm{p}_{\text{o}}\bm{q}_{\text{o}i}^\top+\bm{p}_{\text{c}}\bm{q}_{\text{c}i}^\top
\end{equation}
where $\bm{q}_{\text{o}i}$ and $\bm{q}_{\text{c}i}$ are the $i$th columns of $\bm{Q}_\text{o}$ and $\bm{Q}_\text{c}$, respectively, and $\bm{p}_{\text{o}}$ and $\bm{p}_{\text{c}}$ are the sums of rows of $\bm{P}_\text{o}$ and $\bm{P}_\text{c}$, respectively.

\subsubsection{Algorithm to obtain the system matrices}
Because $\bm{p}_{\text{o}i}$ and $\bm{p}_{\text{c}i}$ are not necessarily orthogonal in our case, we cannot use SVD to decompose $\bm{Y}_{\text{h}i}$ and compute $\bm{p}_{\text{o}i}$, $\bm{p}_{\text{c}i}$, $\bm{q}_{\text{o}}$ and $\bm{q}_{\text{c}}$. Instead, we employ an iterative approach to solve the system matrices. 
From the definition of $\bm{P}_\text{o}$, $\bm{Q}_\text{o}$, the sums of rows of $\bm{P}_\text{o}$ and $\bm{Q}_\text{o}$ nearly maintain a constant value,
as shown in Fig. \ref{fig:t2s_model_e}. 
Conversely, the sums of rows of $\bm{P}_\text{c}$ and $\bm{Q}_\text{c}$ exhibit varying values, as shown in Fig. \ref{fig:t2s_model_f}. This property allows us to generate initial estimates for $\bm{q}_{\text{o}}$ and $\bm{q}_{\text{c}}$, and then use these estimates to progressively solve each column of the system matrices. This process is repeated iteratively to achieve the final calibration results. The full calibration algorithm consists of Algorithm \ref{alg:procedure}, \ref{alg:initial_estimates} and \ref{alg:refine}, which are introduced as follows:

Algorithm \ref{alg:procedure} calculates $\bm{P}_{\text{o}}$ and $\bm{P}_{\text{c}}$ given a list of calibration images $\{\bm{Y}_{i}\}_{i=1}^N$ and $\bm{q}_{\text{o}}$, $\bm{q}_{\text{c}}$. Note that when this algorithm is used to calculate $\bm{P}_{\text{o}}$, $\bm{P}_{\text{c}}$, the input calibration images are the horizontal ones $\{\bm{Y}_{\text{h}i}\}_{i=1}^N$. It can also be used to get $\bm{Q}_{\text{o}}$ and $\bm{Q}_{\text{c}}$, simply by transposing all the vertical calibration images to $\{\bm{Y}_{\text{v}i}^\top\}_{i=1}^M$.
\begin{algorithm}

\renewcommand{\algorithmicrequire}{\textbf{Input:}}
\renewcommand{\algorithmicensure}{\textbf{Output:}}
\caption{get\_Po\_Pc\_given\_qo\_qc}\label{alg:procedure}
\begin{algorithmic}
\REQUIRE list of calibration images $\{\bm{Y}_{i}\}_{i=1}^N$ and two vectors $\bm{q}_{\text{o}}$, $\bm{q}_{\text{c}}$
\ENSURE system matrices $\bm{P}_{\text{o}}$, $\bm{P}_{\text{c}}$
\FOR{$i=1,\dots,n$}
    \STATE $\bm{p}_{\text{o}i}\leftarrow\bm{Y}_{i}\bm{q}_{\text{o}}/(\bm{q}_{\text{o}}^\top\bm{q}_{\text{o}})$
    \STATE $\bm{Y}_{i} \leftarrow \bm{Y}_{i} - \bm{p}_{\text{o}i}\bm{q}_{\text{o}}^\top$
    \STATE $\bm{p}_{\text{c}i}\leftarrow\bm{Y}_{i}\bm{q}_{\text{c}}/(\bm{q}_{\text{c}}^\top\bm{q}_{\text{c}})$
\ENDFOR
\STATE $\bm{P}_{\text{o}} \leftarrow [\bm{p}_{\text{o}1}, \bm{p}_{\text{o}2}, \dots, \bm{p}_{\text{o}n}]$
\STATE $\bm{P}_{\text{c}} \leftarrow [\bm{p}_{\text{c}1}, \bm{p}_{\text{c}2}, \dots, \bm{p}_{\text{c}n}]$
\RETURN $\bm{P}_{\text{o}}$, $\bm{P}_{\text{c}}$

\end{algorithmic}
\end{algorithm}

Then, we define Algorithm \ref{alg:initial_estimates} to generate the initial estimates of the system matrices. We first assume $\bm{q}_{\text{o}}=\bm{1}$ and get the initial $\bm{P}_{\text{o}}$. From the calibration image $\bm{Y}_{\text{h}[N/2]}$, we can get the estimate of $\bm{q}_{\text{c}}$, and then obtain the initial $\bm{P}_{\text{c}}$. We use $\bm{Y}_{\text{h}[N/2]}$ because the slit is positioned at the vertical center of the scene, and therefore its emitted light should mostly be captured by the CMOS sensor, making it easier to estimate $\bm{q}_{\text{c}}$. After $\bm{P}_{\text{o}}$, $\bm{P}_{\text{c}}$ are obtained, we call Algorithm \ref{alg:procedure} to get the initial $\bm{Q}_{\text{o}}$ and $\bm{Q}_{\text{c}}$. 

\begin{algorithm}
\renewcommand{\algorithmicrequire}{\textbf{Input:}}
\renewcommand{\algorithmicensure}{\textbf{Output:}}
\caption{Generate initial estimates of system matrices}\label{alg:initial_estimates}
\begin{algorithmic}
\REQUIRE list of calibration images $\{\bm{Y}_{\text{h}i}\}_{i=1}^N$ and $\{\bm{Y}_{\text{v}i}\}_{i=1}^M$
\ENSURE initial estimates of system matrices $\bm{P}_{\text{o}}$, $\bm{Q}_{\text{o}}$, $\bm{P}_{\text{c}}$, $\bm{Q}_{\text{c}}$

\STATE \textit{First, generate the initial estimate for} $\bm{q}_{\text{o}}$ \textit{and get} $\bm{P}_{\text{o}}$:
\STATE $\bm{q}_{\text{o}}\leftarrow\bm{1}$
\FOR{$i=1,\dots,N$}
    \STATE $\bm{p}_{\text{o}i}\leftarrow\bm{Y}_{\text{h}i}\bm{q}_{\text{o}}/(\bm{q}_{\text{o}}^\top\bm{q}_{\text{o}})=\bm{Y}_{\text{h}i}\cdot\bm{1}/N$
\ENDFOR
\STATE $\bm{P}_{\text{o}}\leftarrow [\bm{p}_{\text{o}1}, \bm{p}_{\text{o}2}, \dots, \bm{p}_{\text{o}N}]$
\STATE \textit{Take the calibration image} $\bm{Y}_{\text{h}[\frac{N}{2}]}$ \textit{to estimate} $\bm{q}_{\text{c}}$:
\STATE $\bm{Y}_{\text{h}[\frac{N}{2}]} \leftarrow \bm{Y}_{\text{h}[\frac{N}{2}]} - \bm{p}_{\text{o}[\frac{N}{2}]}\cdot\bm{1}^\top$
\STATE Perform SVD of $\bm{Y}_{\text{h}[\frac{N}{2}]}$: $\bm{Y}_{\text{h}[\frac{N}{2}]} = \bm{U}\bm{\Sigma}\bm{V}^\top$
\STATE $\bm{q}_{\text{c}}\leftarrow$ the first column of $\bm{V}$
\FOR{$i=1,\dots,N$}
    \STATE $\bm{Y}_{\text{h}i}\leftarrow\bm{Y}_{\text{h}i} - \bm{p}_{\text{o}i}\cdot\bm{1}^\top$
    \STATE $\bm{p}_{\text{c}i}\leftarrow\bm{Y}_{\text{h}i}\bm{q}_{\text{c}}/(\bm{q}_{\text{c}}^\top\bm{q}_{\text{c}})$
\ENDFOR
\STATE $\bm{P}_{\text{c}}\leftarrow [\bm{p}_{\text{c}1}, \bm{p}_{\text{c}2}, \dots, \bm{p}_{\text{c}N}]$
\STATE \textit{Then, get }$\bm{Q}_{\text{o}}$, $\bm{Q}_{\text{c}}$ \textit{based on the initial} $\bm{P}_{\text{o}}$, $\bm{P}_{\text{c}}$:
\STATE $\bm{p}_\text{o}\leftarrow\bm{P}_{\text{o}}\cdot\bm{1}$, $\bm{p}_\text{c}\leftarrow\bm{P}_{\text{c}}\cdot\bm{1}$
\STATE $\bm{Q}_{\text{o}}$, $\bm{Q}_{\text{c}}$ $\leftarrow$ get\_Po\_Pc\_given\_qo\_qc($\{\bm{Y}_{\text{v}i}^\top\}_{i=1}^M$, $\bm{p}_\text{o}$, $\bm{p}_\text{c}$) (Alg.~\ref{alg:procedure})

\RETURN $\bm{P}_{\text{o}}$, $\bm{P}_{\text{c}}$, $\bm{Q}_{\text{o}}$, $\bm{Q}_{\text{c}}$,

\end{algorithmic}
\end{algorithm}

Once the initial estimates of all system matrices are ready, we can progressively refine the system matrices, as shown in Algorithm \ref{alg:refine}. In each iteration, the system matrices are scaled to have equal Frobenius norms to avoid divergence.
\begin{algorithm}
\renewcommand{\algorithmicrequire}{\textbf{Input:}}
\renewcommand{\algorithmicensure}{\textbf{Output:}}
\caption{Refine the system matrices}\label{alg:refine}
\begin{algorithmic}
\REQUIRE initial estimates of system matrices $\bm{P}_{\text{o}}$, $\bm{Q}_{\text{o}}$, $\bm{P}_{\text{c}}$, $\bm{Q}_{\text{c}}$
\ENSURE refined system matrices $\bm{P}_{\text{o}}$, $\bm{Q}_{\text{o}}$, $\bm{P}_{\text{c}}$, $\bm{Q}_{\text{c}}$

\REPEAT
\STATE $\bm{q}_\text{o}\leftarrow\bm{Q}_{\text{o}}\cdot\bm{1}$, $\bm{q}_\text{c}\leftarrow\bm{Q}_{\text{c}}\cdot\bm{1}$
\STATE $\bm{P}_{\text{o}}$, $\bm{P}_{\text{c}}$ $\leftarrow $get\_Po\_Pc\_given\_qo\_qc($\{\bm{Y}_{\text{h}i}\}_{i=1}^N$, $\bm{q}_\text{o}$, $\bm{q}_\text{c}$) (Alg.~\ref{alg:procedure})
\STATE $\bm{p}_\text{o}\leftarrow\bm{P}_{\text{o}}\cdot\bm{1}$, $\bm{p}_\text{c}\leftarrow\bm{P}_{\text{c}}\cdot\bm{1}$
\STATE $\bm{Q}_{\text{o}}$, $\bm{Q}_{\text{c}}$ $\leftarrow $get\_Po\_Pc\_given\_qo\_qc($\{\bm{Y}_{\text{v}i}^\top\}_{i=1}^M$, $\bm{p}_\text{o}$, $\bm{p}_\text{c}$) (Alg.~\ref{alg:procedure})
\STATE Scale $\bm{P}_{\text{o}}$, $\bm{Q}_{\text{o}}$ to ensure $\Vert\bm{P}_{\text{o}}\Vert_\text{F}=\Vert\bm{Q}_{\text{o}}\Vert_\text{F}$
\STATE Scale $\bm{P}_{\text{c}}$, $\bm{Q}_{\text{c}}$ to ensure $\Vert\bm{P}_{\text{c}}\Vert_\text{F}=\Vert\bm{Q}_{\text{c}}\Vert_\text{F}$
\UNTIL{convergence or max number of iterations reached}
\RETURN $\bm{P}_{\text{o}}$, $\bm{P}_{\text{c}}$, $\bm{Q}_{\text{o}}$, $\bm{Q}_{\text{c}}$,

\end{algorithmic}
\end{algorithm}

\subsection{Interpretation of Tactile Information}
\label{sec:depth_recon}
In this subsection, we introduce the method for reconstructing the contact geometry (i.e., the depth map) from the reconstructed tactile images. Further, we adopt a Real2Sim method to transfer the real tactile images into the simulation domain to improve the depth reconstruction quality. Finally, we introduce the method for interpreting marker displacements.

\noindent \textbf{Reconstruction of contact geometry.}
In the proposed design of the tactile sensor, the contact geometry is reconstructed using the photometric stereo method~\cite{johnson2009retrographic,yuan2017gelsight}. The key idea of photometric stereo is to map the surface gradient into color intensities, and then the depth map can be computed using the Poisson equation. Typically, the mapping from color intensities to gradient values can be modelled in two ways: using a Look-Up Table (LUT)~\cite{yuan2017gelsight} or using a neural network~\cite{li2018end,wang2021gelsight}. In this work, we adopt the LUT method.

\noindent \textbf{Real2Sim method for improving depth reconstruction.}
In real sensors, several non-ideal properties can impact the accuracy of photometric stereo, such as non-uniform illumination, inter-reflections, shadows and artifacts existing in lensless reconstruction. To address these issues, we employ a Real2Sim transfer method based on CycleGAN~\cite{zhu2017unpaired} to mitigate the non-ideal factors present in reality. This method has been validated on standard GelSight sensors in our previous work \cite{chen2022bidirectional}. In the experiment section, we will demonstrate that this method is also applicable to the proposed tactile sensor.

\noindent \textbf{Interpreting marker displacements.}
Markers are frequently employed in vision-based tactile sensors to detect shear deformation and forces. In order to interpret the displacements of these markers, we establish a correspondence between consecutive frames by applying marker detection and the nearest-neighbor algorithm. By cascading this correspondence, we can obtain the current marker displacements relative to the initial reference frame. The displacement fields derived from the markers can then be leveraged for manipulation tasks.

\section{Experiment: Sensor Fabrication}\label{sec:experiments_fabrication}

In Sec. \ref{sec:methods} and Fig. \ref{fig:work_flow}, we present the conceptual design of \name. In this section, the detailed fabrication process of \name~will be introduced.

\begin{figure}[htbp]
    \centering
\includegraphics[width=7.5cm]{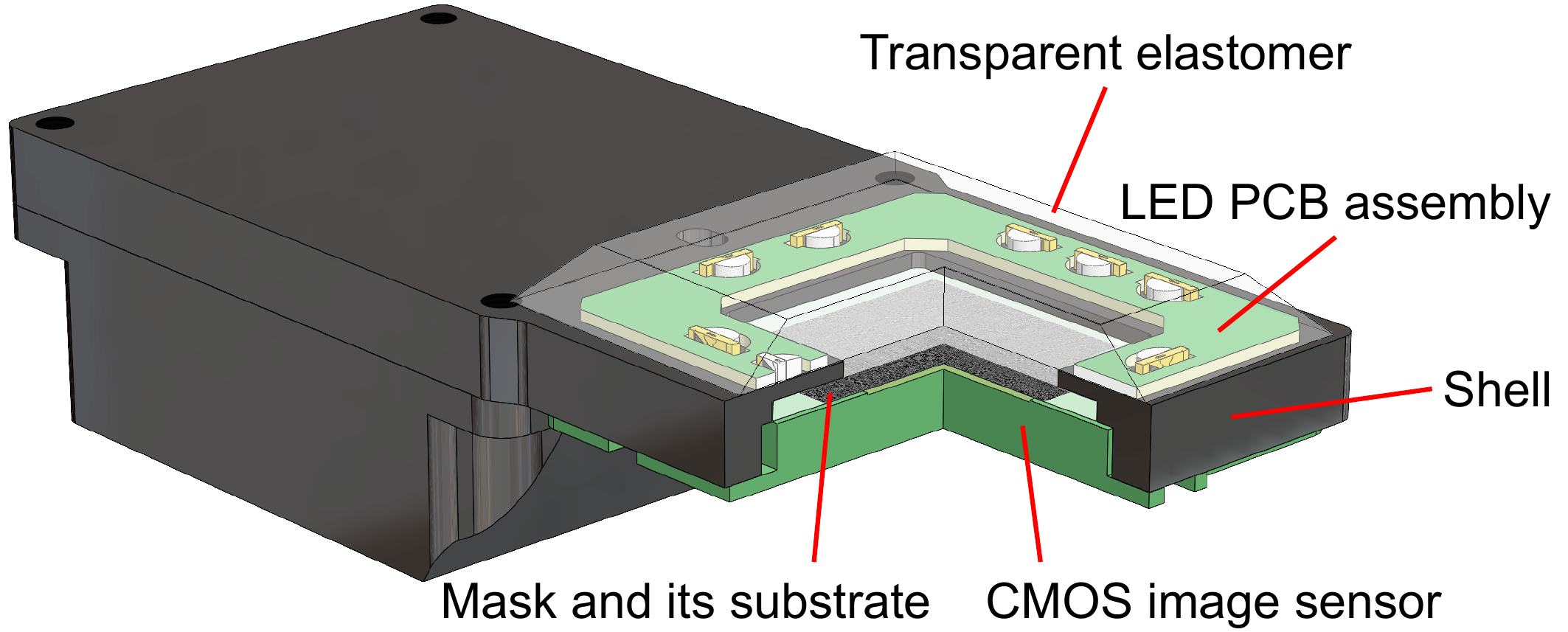}
    \caption{CAD model of \name. This figure mainly presents the lensless imaging subsystem, while the reflective membrane is not shown.}
    \label{fig:lensless_image_component}
\end{figure}

Fig. \ref{fig:lensless_image_component} presents the CAD model of \name~without the reflective membrane. To fabricate \name, we start with constructing the lensless imaging subsystem. It comprises the following steps:

\begin{enumerate}[(1)] 
\item \textit{Prepare the sensor shell.} The shell is made of aluminum alloy through CNC machining, which is shown in black in Fig. \ref{fig:lensless_image_component}.

\item \textit{Place the separable amplitude mask.} The mask pattern is created using a thin chrome film on a 1-mm-thick piece of soda glass. The feature size of the mask $\delta_\text{mask}$, which equals the unit size of the mask pattern, is 20 \textmu m.

\item \textit{Install the CMOS sensor.} Beneath the mask is the CMOS sensor, which is fixed to the sensor shell by screws. We have modified a JAI GO-5000C-USB industrial camera to construct our lensless camera. The GO-5000C-USB camera has a 1" optical format and a resolution of 2560$\times$2048. To reduce the amount of data, we do not perform demosaicing on the raw image. Instead, we extract the red, blue pixels respectively, and average the two green pixels in each Bayer filter unit to form a 3-channel, 1280$\times$1024 image.

\item \textit{Attach the LED PCB assembly.} The LEDs provide illumination of different colors from different directions, which is required by Photometric Stereo. We use 10 side-illuminating LEDs and place them around the mask, as shown in Fig. \ref{fig:lensless_image_component}. Their colors are red, green, blue, and white, respectively.

\item \textit{Cast the transparent elastomer.} Following DIGIT \cite{lambeta2020digit}, we use Smooth-On Solaris silicone rubber as the elastomer's material. The elastomer is directly cast onto the shell, the soda glass and the LED PCB assembly, using a CNC-machined aluminum mold. The thickness of the elastomer is 3 mm.

\end{enumerate}

After constructing the lensless imaging subsystem, we perform calibration to obtain the system matrices and conduct experiments to validate its performance, which will be presented in Sec. \ref{sec:experiments_lensless_imaging_subsystem}. Later, we complete \name's fabrication by sequentially applying thin layers of opaque elastomer and laser-cutting the markers. The details are as follows:

\begin{figure}
    \centering
    \subfigure[]{
		\includegraphics[width=7cm]{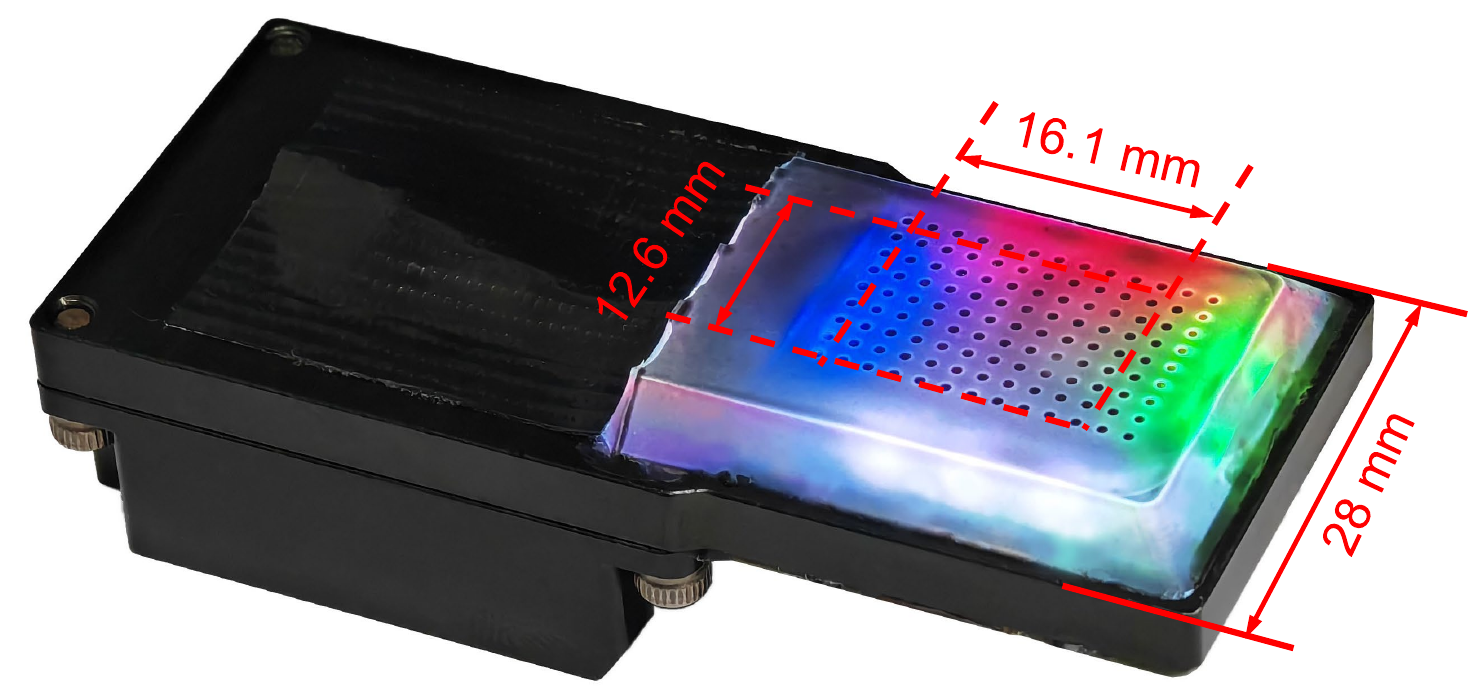}
		\label{fig:tactile_sensor_real_a}
	}
     \subfigure[]{
		\includegraphics[width=5cm]{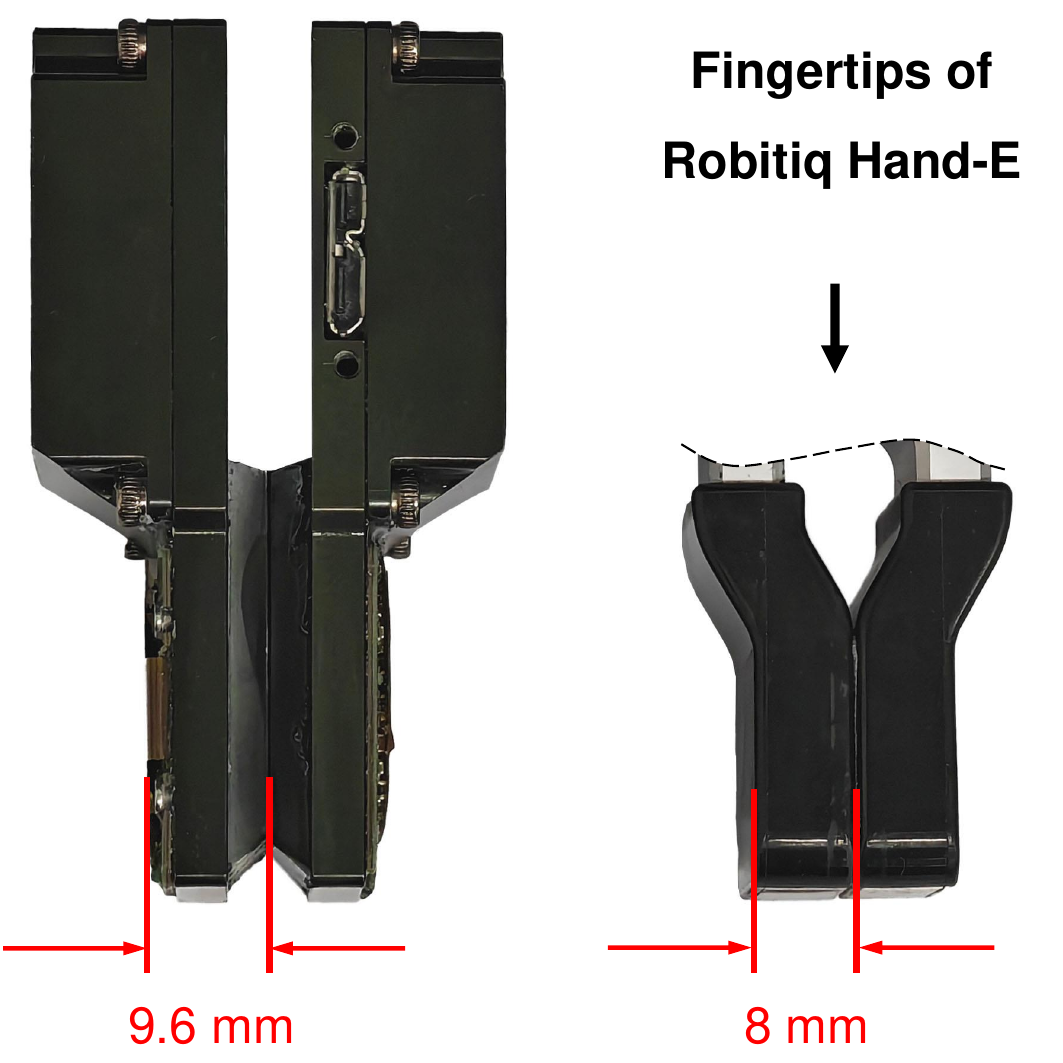}
		\label{fig:tactile_sensor_real_b}
	}
    \caption{Dimensions of \name. (a) \name~with all its LEDs lighted. Its sensing field is 16.1$\times$12.6 mm$^2$ as indicated by the red dashed lines. (b) Comparison of \name~and the fingertips of Robotiq Hand-E. While offering high-resolution tactile sensing, \name~is 9.6 mm thick and is only slightly thicker than typical fingertips.}
    \label{fig:tactile_sensor_real}
\end{figure}

\begin{enumerate}[(1)] 
\item \textit{Prepare white and black paint mixtures.} We follow the process of DIGIT~\cite{lambeta2020digit} for this step. Firstly, we mix an equal part A/B solution of Smooth-On EcoFlex 00-10. Then, we add Smooth-On White (Black) Silicone Pigment such that it is 3\% of the total weight of the previous mixture. Finally, we thin out the mixture to 20\% of the total weight using Smooth-On NOVOCS Matte.

\item \textit{Apply the white paint mixture.} We use an air brush to spray the white paint mixture onto the transparent elastomer's surface and form a thin layer.

\item \textit{Laser-cut the markers.} We use a laser-cutting machine to form small holes on the white layer. The diameter of the holes is 0.5 mm. The holes have a 1.5 mm spacing.
\item \textit{Apply the black paint mixture.} We use another air brush to spray the black paint mixture onto the surface. It provides the black color for the markers and eliminates the influence of environmental light.

\end{enumerate}

After all the fabrication process is completed, the sensor is shown in Fig. \ref{fig:tactile_sensor_real_a}. Fig. \ref{fig:tactile_sensor_real_b} presents the thickness of \name~and compares it with the fingertips of Robotiq Hand-E. Despite providing a sensing field of over 200 mm$^2$, \name~maintains a slim profile. At only 9.6 mm thick, it is slightly thicker than a typical robotic fingertip and thinner than human fingers.

\section{Experiment: Lensless Imaging Subsystem}\label{sec:experiments_lensless_imaging_subsystem}

\subsection{Configuration of \name's Lensless Imaging Subsystem}\label{sec:experiment_lensless_configuration}

\begin{figure}[htbp]
    \centering
\includegraphics[width=5cm]{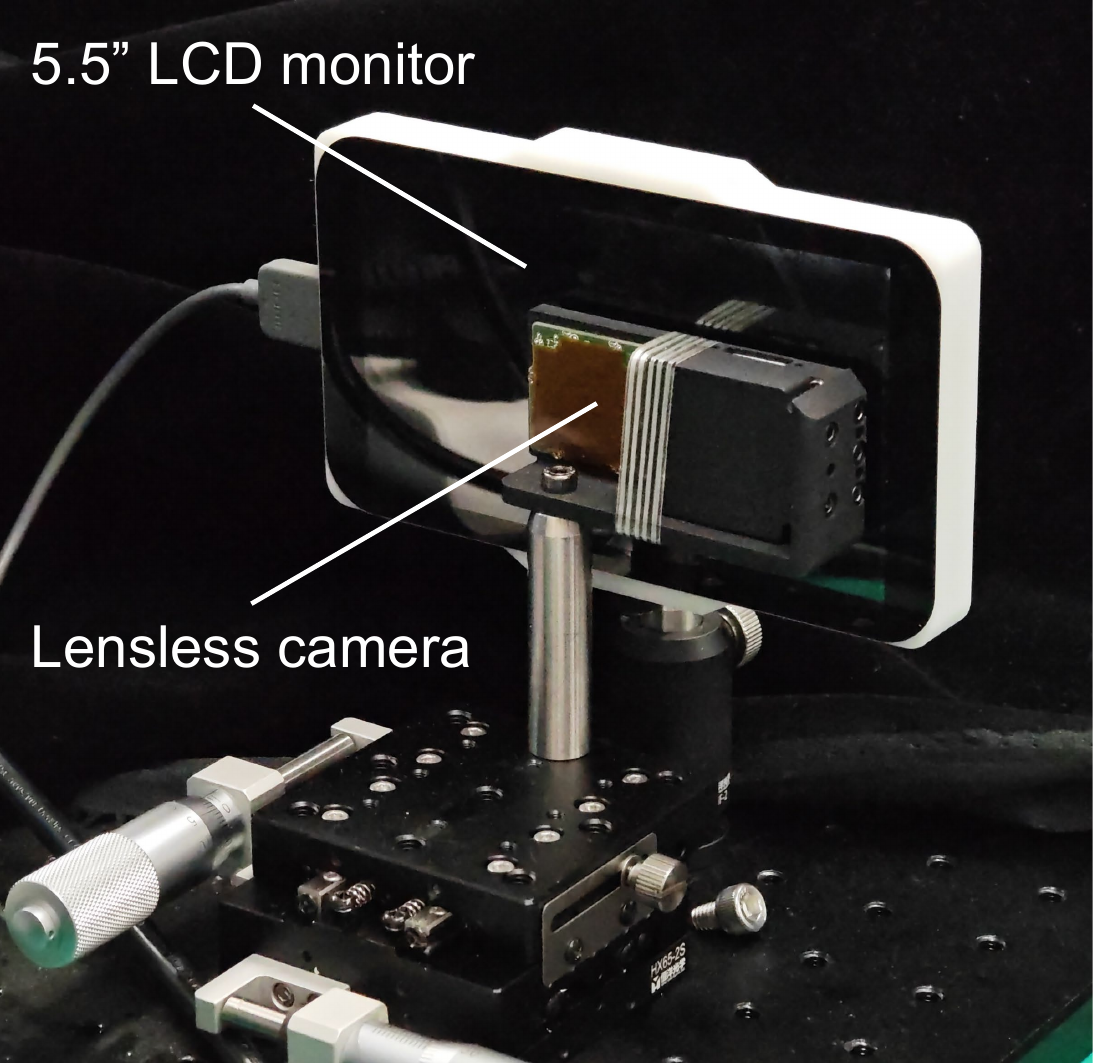}
    \caption{Calibration setup for the lensless imaging subsystem. We use a 5.5" LCD monitor to display the slit images for calibration.}
    \label{fig:calibration_setup}
\end{figure}

The fabrication of the lensless imaging subsystem has been introduced in Sec. \ref{sec:experiments_fabrication}. Lensless imaging experiments are conducted before the transparent elastomer is covered with the refractive membrane. The calibration setup is shown in Fig. \ref{fig:calibration_setup}. The calibration images are displayed using a 5.5" 3840$\times$2160 LCD monitor, which is tightly attached to the transparent elastomer. The pixel size of the monitor is 31.5 \textmu m, and therefore the pixel size of the reconstructed image $\delta_\text{scene}$ is also 31.5 \textmu m. The resolution of the reconstructed images is determined to be 512$\times$400, which corresponds to an area of 16.1$\times$12.6 mm$^2$.
The key technical specifications of the lensless imaging system are summarized in Table \ref{tab:lensless_parameters}.

\begin{figure*}
    \centering
    \includegraphics[width=18cm]{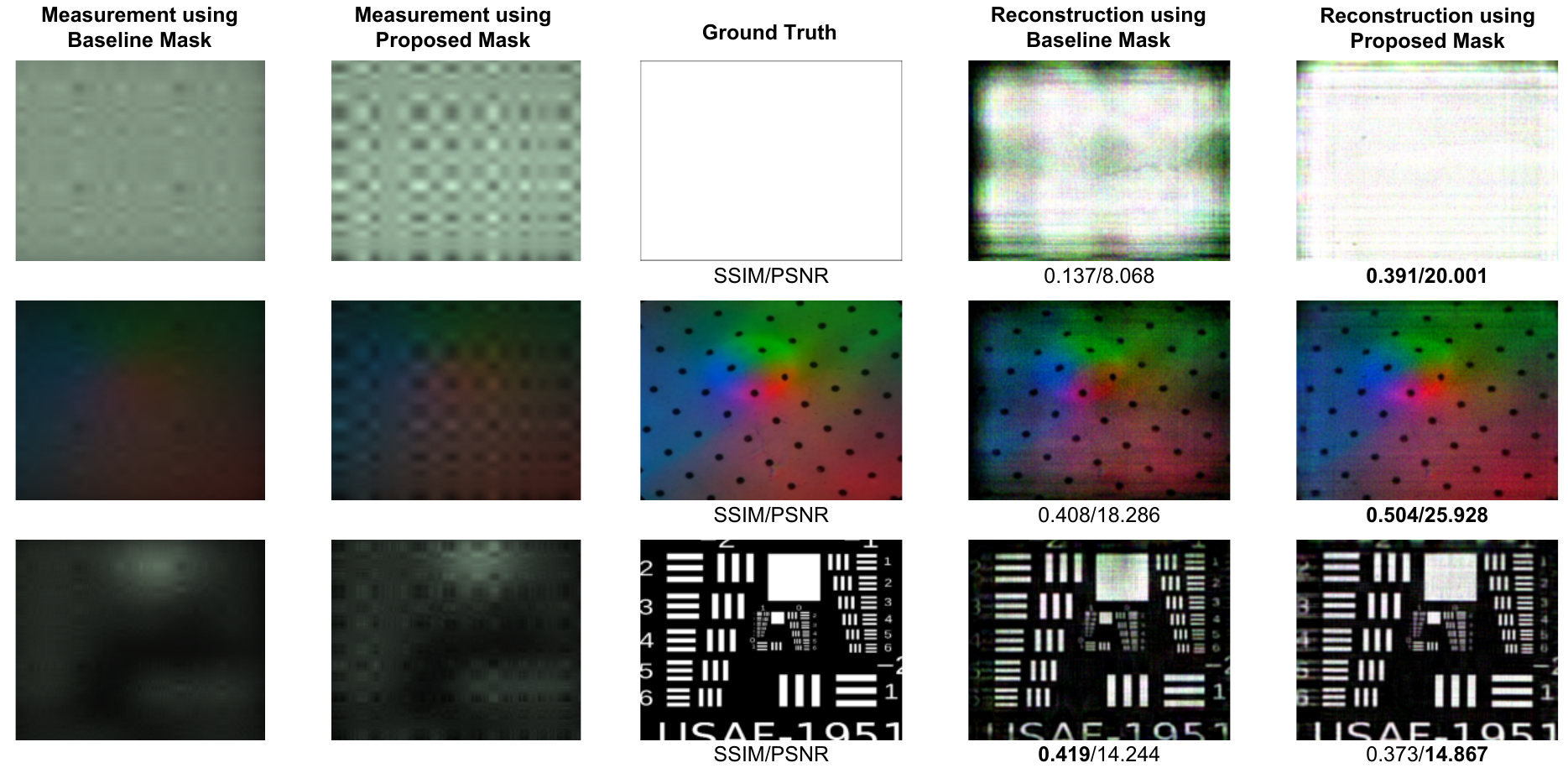}
    \caption{Comparison of the reconstruction qualities using the proposed optimized mask and the traditional MLS-based mask. The reconstruction algorithm is the proposed non-iterative algorithm. Three representative scenes are tested. The first row is a pure white image for testing the uniformity of reconstruction. The second row is the tactile image from a traditional GelSight sensor. The third row is the reshaped USAF-1951 resolution test chart. From the pure white scene, one can see that the uniformity of reconstructions is greatly improved using the optimized mask. The quantitative metrics also prove the superiority of the proposed method.}
    \label{fig:compare_mask}
\end{figure*}

\begin{table}[h]
    \caption{Key Technical Specifications of the Lensless Imaging System}
    \centering
    \begin{tabular}{lc}
    \toprule
        Name & Value\\
    \midrule
         CMOS resolution & 2560$\times$2048\\
         CMOS pixel size & 5 \textmu m\\
         CMOS size & 12.8 mm$\times$10.24 mm\\
         measurement resolution $S\times R$& 1280$\times$1024\\
         effective pixel size of measurement $\delta_\text{sensor}$& 10 \textmu m\\
         reconstruction resolution $M\times N$ & 512$\times$400\\
         pixel size of reconstruction $\delta_\text{scene}$& 31.5 \textmu m\\
         area size of reconstruction & 16.1 mm$\times$12.6 mm\\
         mask feature size $\delta_\text{mask}$& 20 \textmu m\\
         scene-to-mask distance $z$ (estimated) & 4.8 mm\\
         mask-to-CMOS distance $d$ (estimated) & 1 mm\\
    \bottomrule
    \end{tabular}
    \label{tab:lensless_parameters}
\end{table}

\subsection{Comparison of the Optimized Mask and the MLS-based Mask}

\begin{figure}[htbp]
    \centering
    \subfigure[]{
    \includegraphics[width=4cm]{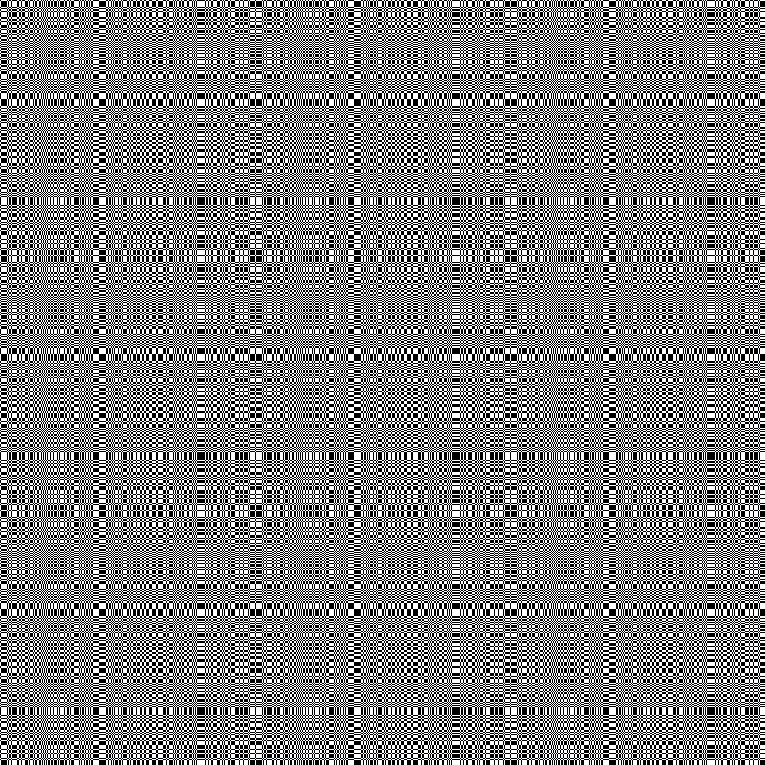}
    \label{fig:mls_based_mask}
    }
    \hfill
    \subfigure[]{
    \includegraphics[width=4cm]{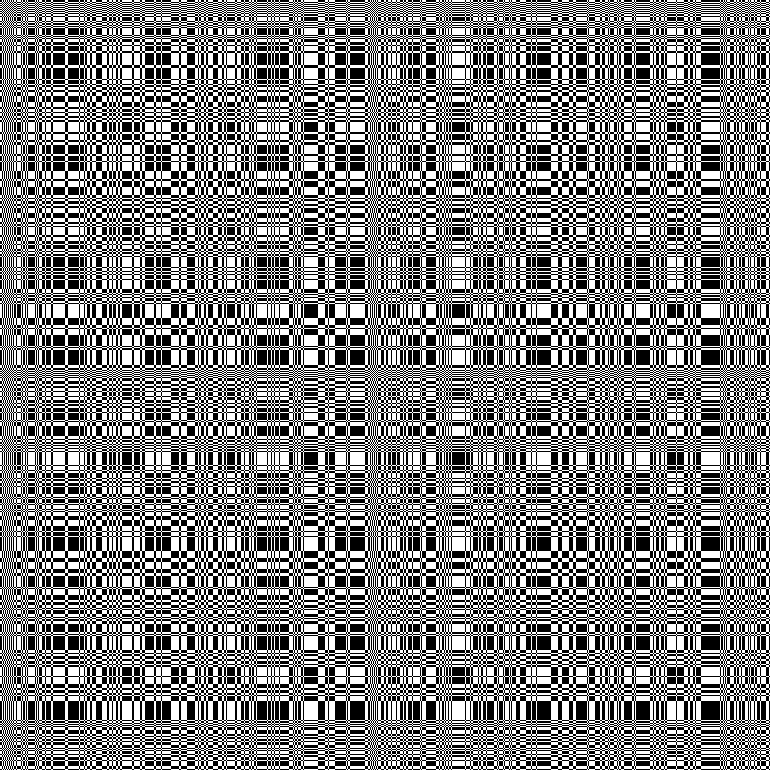}
    \label{fig:optimized_mask}
    }
    
    \caption{Comparison of (a) the MLS-based mask and (b) the optimized mask obtained from the genetic algorithm.}
    \label{fig:mls_and_optimized_mask}
\end{figure}

\begin{figure*}
    \centering
    \includegraphics[width=16cm]{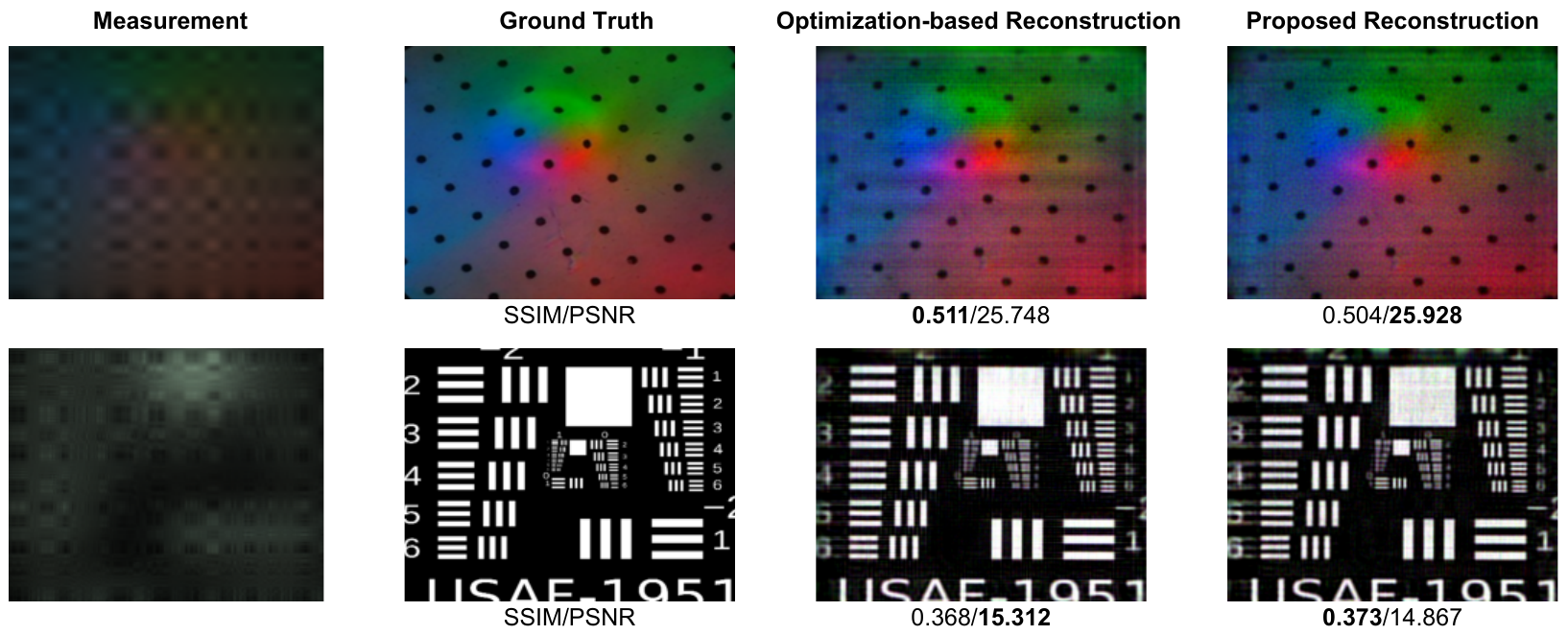}
    \caption{Comparison of the proposed non-iterative reconstruction algorithm and the traditional optimization-based algorithm~\cite{adams2017single}. Two representative scenes are tested. The first row is the tactile image from a standard GelSight sensor and the second row is the reshaped USAF-1951 resolution test chart. The proposed algorithm exhibits much higher efficiency while remaining similar reconstruction qualities as the traditional one.}
    \label{fig:compare_reconstruction}
\end{figure*}

To generate an optimized mask, the vector $\bm{\varphi}$'s length $K$ is set to 770. We run the genetic algorithm with 2000 generations and 200 individuals in each generation. The crossover and mutation probability are set to be 0.8 and 0.1, respectively. In each mutation operation, 10 random loci in $\bm{\varphi}$ are inverted. To evaluate each individual's fitness, we choose the pure white image $\bm{X}_\text{white}$~(the 1st row in Fig. \ref{fig:compare_mask}) and the reshaped USAF test image $\bm{X}_\text{USAF}$~(the 3rd row in Fig. \ref{fig:compare_mask}), and the fitness score of gene $\bm{\varphi}$ is calculated as:
\begin{equation}
\begin{aligned}
    fitness(\bm{\varphi})&=\frac{2f_\text{SSIM}(\bm{\varphi},\bm{X}_\text{USAF})f_\text{SSIM}(\bm{\varphi},\bm{X}_\text{white})}{f_\text{SSIM}(\bm{\varphi},\bm{X}_\text{USAF})+f_\text{SSIM}(\bm{\varphi},\bm{X}_\text{white})}\\&+
    \frac{f_\text{PSNR}(\bm{\varphi},\bm{X}_\text{USAF})+f_\text{PSNR}(\bm{\varphi},\bm{X}_\text{white})}{2}\\&+
    f_\text{GRAD}(\bm{\varphi},\bm{X}_\text{white})
\end{aligned}
\end{equation}
Finally, the individual with the highest fitness score is selected. The corresponding optimized mask pattern is shown in Fig. \ref{fig:optimized_mask}. It covers an area of 15.4$\times$15.4 mm$^2$, sufficient to overlay the full extent of the CMOS underneath.

For comparison, we choose the MLS-based mask in \cite{asif2016flatcam} as the baseline. We use a 255-length MLS and repeat it 3 times to be a 765-length sequence. Then, we use \eqref{eq:mask_generation} to form a 15.3$\times$15.3 mm$^2$ mask (Fig. \ref{fig:mls_based_mask}), as the baseline.

The experimental setup is the same as the calibration setup introduced in \ref{sec:experiment_lensless_configuration}, as shown in Fig. \ref{fig:calibration_setup}. The ground-truth images are displayed on the monitor, and the measurement images are obtained by averaging 100 frames from the CMOS. SSIM and PSNR are used as the quantitative metrics to evaluate the reconstruction quality.
 
In Fig. \ref{fig:compare_mask}, we present the comparison results. Fig. \ref{fig:compare_mask} presents three representative scenes. From qualitative comparison and quantitative metrics, we draw the conclusion that our proposed algorithm has largely improved the reconstruction qualities compared with the MLS-based mask, especially in terms of uniformity. Another significant advantage is the larger usable FOV provided by the optimized mask. The reason is as follows: In lensless imaging, a point at the FOV boundary only casts a small PSF area on the CMOS. With the MLS-based mask, the CMOS collects minimal information about these boundary points, resulting in decaying intensities in the reconstructed images's boundary due to the regularization term in \eqref{eq:optimization_tik}. In contrast, the optimized mask takes this factor into account, leading to a larger usable FOV. These advantages will considerably enhance subsequent tactile sensing.

\subsection{Comparison of Lensless Reconstruction Algorithms}

A reconstruction algorithm with high computational efficiency and good quality is crucial in the application of tactile sensing. Therefore, we conduct experiments to compare the proposed non-iterative reconstruction and the traditional optimization-based reconstruction algorithm. Specifically, for the optimization-based reconstruction algorithm, we use Nesterov gradient method to solve \eqref{eq:T2S_optimization}.

In Fig. \ref{fig:compare_reconstruction}, we present the comparison results of two representative scenes. From qualitative comparison and quantitative metrics, we draw the conclusion that our proposed algorithm has reconstruction qualities similar to the traditional optimization-based method.

We further compare the reconstruction speed by computing the average time of reconstructing 100 images. For the optimization-based method, we have tuned the parameters so that it only takes 800 iterations to converge. Both algorithms are implemented in PyTorch. We used a NVIDIA RTX 3080 Ti GPU and an Intel Core i7-13700K CPU for the experiment. Table~\ref{tab:compare_speed} summarizes the results, which clearly reveals that the proposed algorithm is about 3 orders of magnitude faster than the optimization-based counterpart. 
Our algorithm can achieve real-time reconstruction up to hundreds of frames per second on a desktop GPU,  providing enough efficiency for most robotic tactile applications.

\begin{table}[h]
    \caption{Comparison of Time Consumption of the Proposed Lensless Reconstruction Algorithm and the Optimization-based One}
    \centering
    \begin{tabular}{lccc}
    \toprule
        Reconstruction algorithm & CPU time (s) & GPU time (s)\\
    \midrule
        The optimization-based method~\cite{adams2017single} & 46.1 & 2.01\\
        The proposed non-iterative method & \textbf{0.0585} & \textbf{0.00150}\\
    \bottomrule
    \end{tabular}
    \label{tab:compare_speed}
\end{table}

\section{Experiment: Tactile Sensor Performance}\label{sec:experiment_tactile_sensor}

In this section, we first compare the compactness of \name~with several typical vision-based tactile sensors with straight optical paths in Sec.\ref{sec:compactness}. Sec. \ref{sec:lateral_resolution} discusses the sensor's lateral resolution. \name~utilizes photometric stereo with Real2Sim for depth reconstruction, which is presented in Sec. \ref{sec:depth_accuracy}. 
As an application of the high-resolution images, we showcase \name's capability to recognize fine fabric textures from a single static touch in Sec. \ref{sec:texture_recognition}.

\subsection{Compactness Comparison}
\label{sec:compactness}
\begin{table*}[htbp]
  \centering
  \caption{Comparison of Typical Vision-based Tactile Sensors With Straight Optical Paths}
      \begin{threeparttable}

    \begin{tabular}{lcccccc}
    \toprule
     Imaging principle & \multicolumn{4}{c}{Lens-based} &Micro-lens-based & Lensless \\
    \midrule
     Sensor name & GelSight Mini~\cite{gelsight2023gelsightmini} & GelSlim3.0~\cite{taylor2022gelslim} & DIGIT~\cite{lambeta2020digit} & DTact~\cite{lin2023dtact} & Chen \etal~\cite{chen2022thin} &{\name~(ours)} \\
    \midrule
    Sensing field $A$ (mm$^2$) & 18.6×14.3 &27×25 & 19×16 & 24×24 & 9×8 & 16.1×12.6 \\
    \midrule
    Thickness $D$ (mm) & 28    & 20    & 28\tnote{*}    & 45\tnote{*}    & \textbf{5}     & \textbf{9.6} \\
    \midrule
    Scene-to-CMOS distance $D_2$ (mm) & 21    & 17\tnote{*}    & 18\tnote{*}    & 40\tnote{*}    & \textbf{4}\tnote{**}     & \textbf{5.8} \\
    \midrule
    $A/D^2$  & 0.34  & 1.69  & 0.39  & 0.28  & \textbf{2.88}  & \textbf{2.20}  \\
    \midrule
    $A/D_2^2$ & 0.60  & 2.34  & 0.94  & 0.36  & 4.50  & \textbf{6.03}  \\
    \bottomrule
    \end{tabular}%
    \end{threeparttable}
    \begin{tablenotes}
        \item * measured from CAD models
        \item ** estimated by assuming its CMOS PCB assembly is 1 mm thick
      \end{tablenotes}
  \label{tab:compare_sensors}%
\end{table*}%

In Table \ref{tab:compare_sensors}, we present a comparison of several typical vision-based tactile sensors with straight optical paths. We introduce a dimensionless metric, which is the sensing area divided by the square of the thickness, to quantify the compactness of the sensor. The table considers two types of thickness: the actual physical thickness of the sensor and the scene-to-CMOS distance which signifies the minimal possible thickness. The table reveals that the majority of vision-based tactile sensors possess a thickness exceeding 20 mm. However, \name~and the MLA-based sensor manage to maintain a thickness of less than 10 mm. In terms of the compactness metric, both \name~and the MLA-based sensor significantly outperform the others. It’s important to note that the CMOS we utilized has a thickness of over 4 mm, which contributes to the relatively larger physical thickness of \name. Nonetheless, when considering the scene-to-CMOS distance, \name~still holds an advantage over the MLA-based sensor.

\subsection{Lateral Resolution}\label{sec:lateral_resolution}
Vision-based tactile sensors are widely recognized for their superior resolution in comparison to tactile sensors using other sensing principles. Typically, the resolution referred to in this context is the lateral resolution, which bears a close relation to the high resolution of the CMOS image sensor. In this study, tactile images are reconstructed from raw measurements. It’s important to note that the actual resolution might differ from the resolution of the reconstructed images. In this subsection, we characterize \name's lateral resolution from two aspects.

\begin{figure}
    \centering
    \subfigure[]{
		\includegraphics[width=8.5cm]{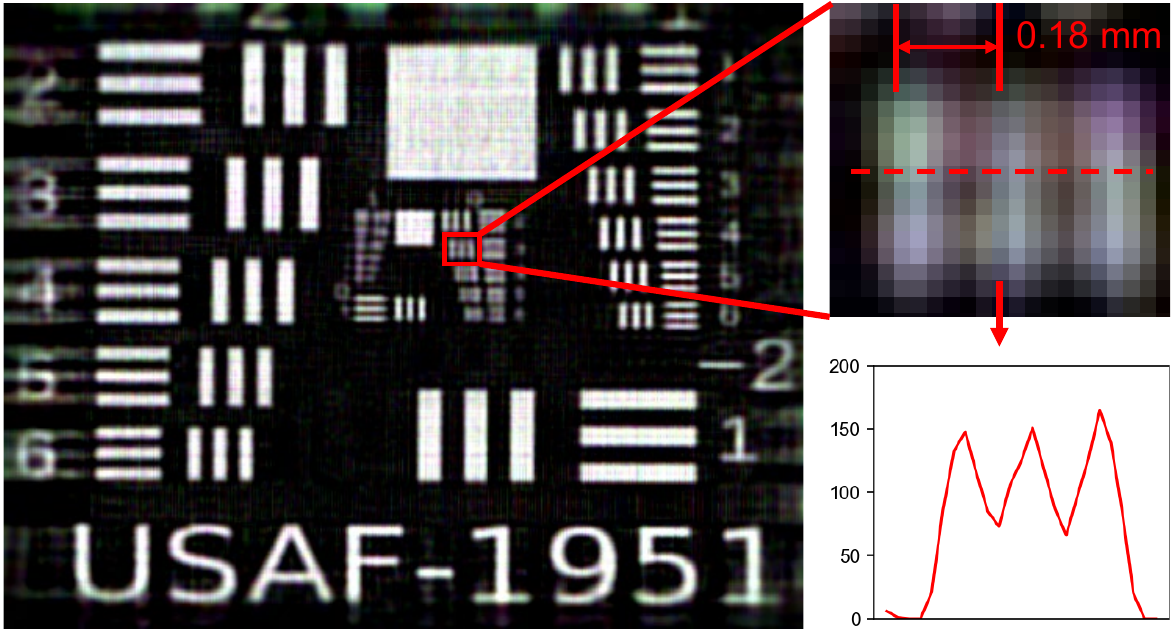}
		\label{fig:resolution_a}
	}

     \subfigure[]{
		\includegraphics[height=2.22cm]{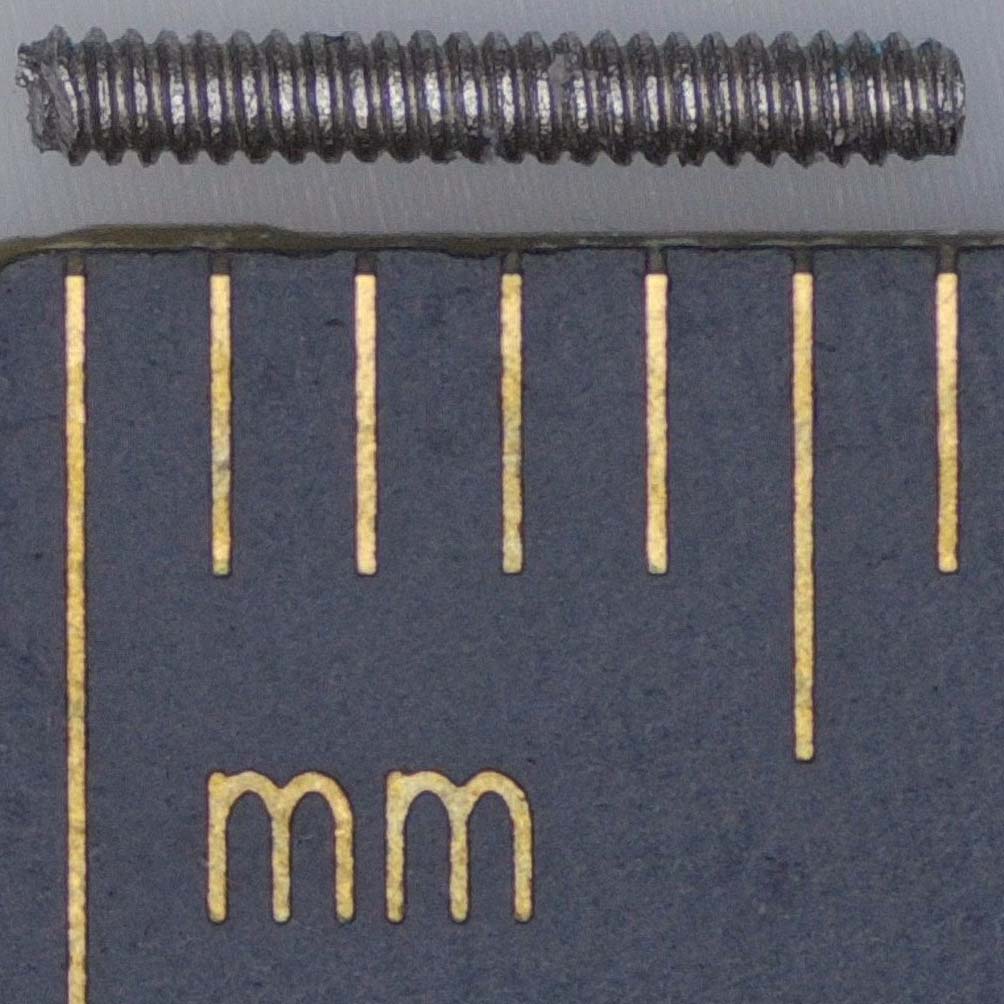}
		\label{fig:screw_microscope}
	}
    \hfill
    \subfigure[]{
		\includegraphics[height=2.22cm]{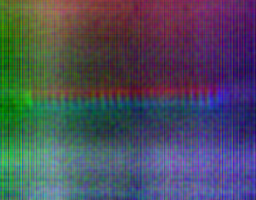}
		\label{fig:screw_tactile}
	}
    \hfill
    \subfigure[]{
    \includegraphics[height=2.22cm]{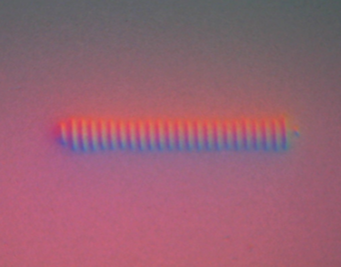}
		\label{fig:screw_gelsightmini}
    }
    
    \caption{Resolution test of \name. (a) Optical resolution. (b) Screw for testing the tactile resolution. (c) Reconstructed tactile image of the screw. (d) GelSight Mini's tactile image of the screw.}
    \label{fig:resolution}
\end{figure}

First, we figure out the optical resolution from the standard resolution test chart. As shown in Fig. \ref{fig:resolution_a}, the minimum spacing of white rectangles that \name~can recognize is about 0.18 mm.
Second, we measure the actual tactile resolution by testing its ability to identify small depth fluctuations. As Shown in Fig. \ref{fig:screw_tactile}, a screw is pressed against the sensor and its screw thread can be clearly seen. The screw's major diameter is 1 mm and its pitch is 0.25 mm (Fig. \ref{fig:screw_microscope}).
The reason that the tactile resolution is lower than the optical resolution is due to the presence of the reflect membrane on the sensor surface, which degrades the transmission and detection of fine geometry details. Fig.~\ref{fig:screw_gelsightmini} shows the tactile image from GelSight Mini~\cite{gelsight2023gelsightmini}, which has a higher resolution than our tactile sensor. This can be attributed to not only the lens system, but also higher camera resolution, softer elastomer, better illumination, and thinner reflect membrane.

\subsection{Depth Accuracy}\label{sec:depth_accuracy}

\begin{figure}
    \centering
    \subfigure[]{
		\includegraphics[width=5cm]{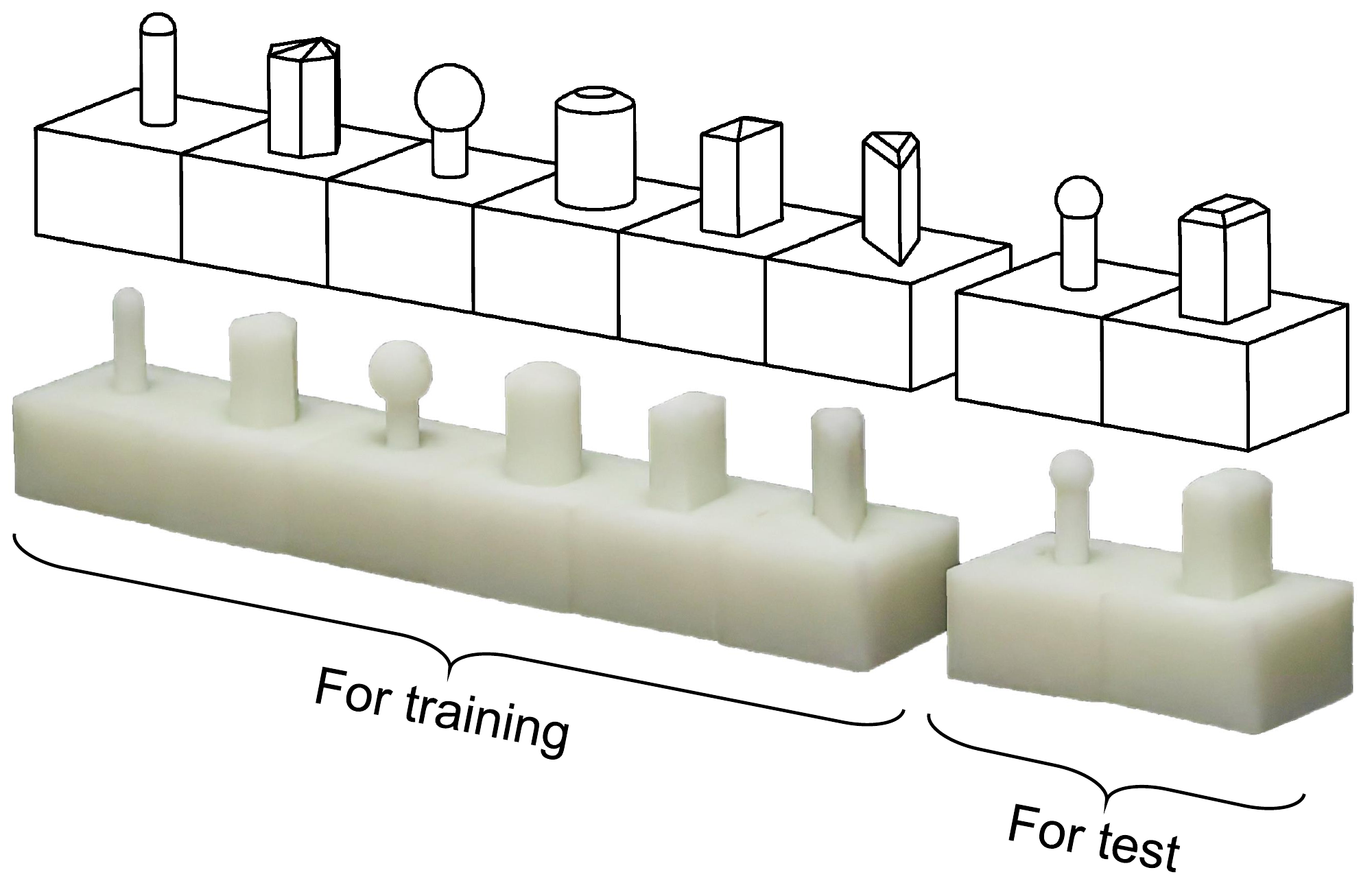}
		\label{fig:indenters}
	}
  \hfill
     \subfigure[]{
		\includegraphics[width=3cm]{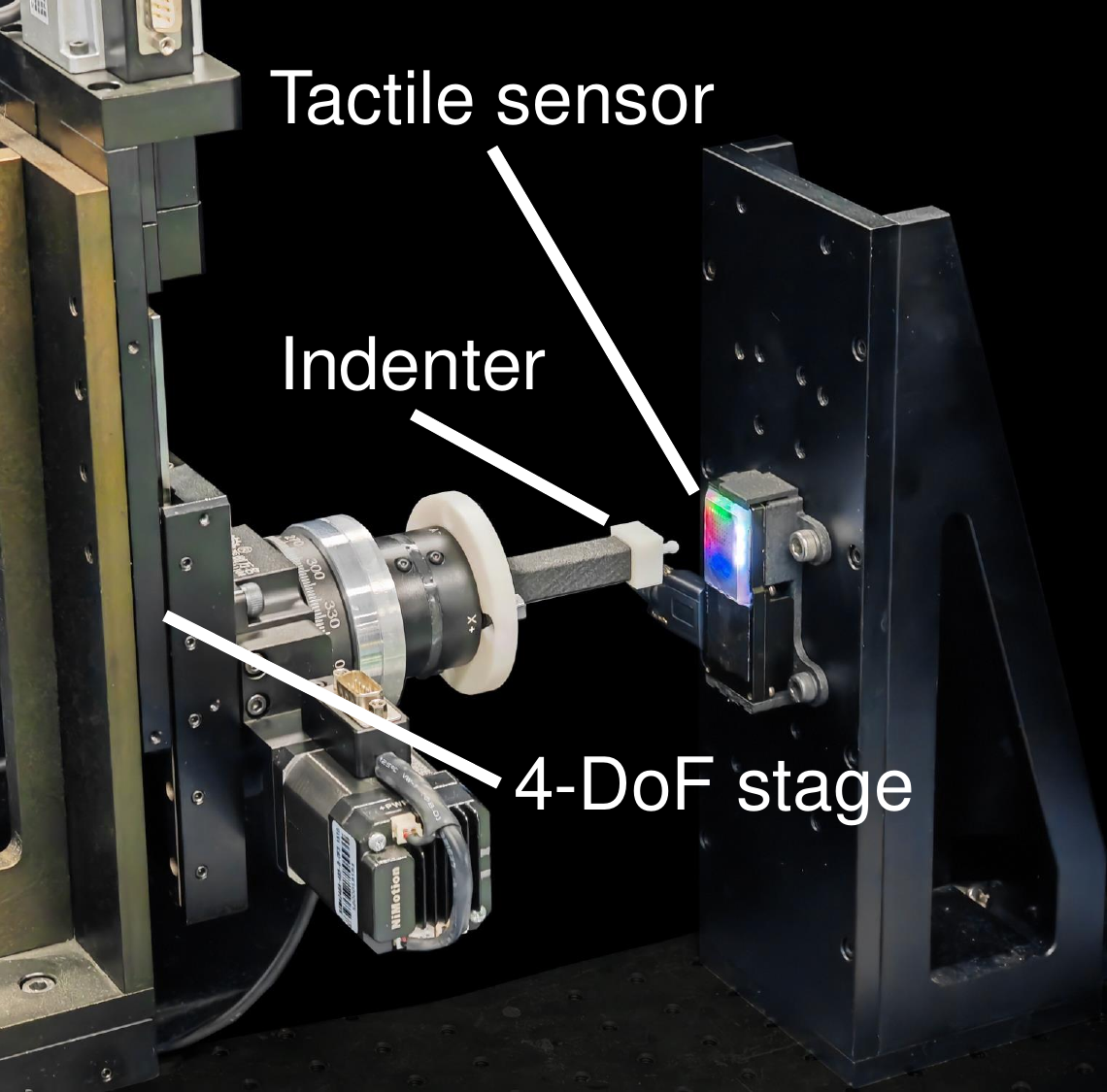}
		\label{fig:indentation_setup}
	}
     \subfigure[]{
		\includegraphics[width=7cm]{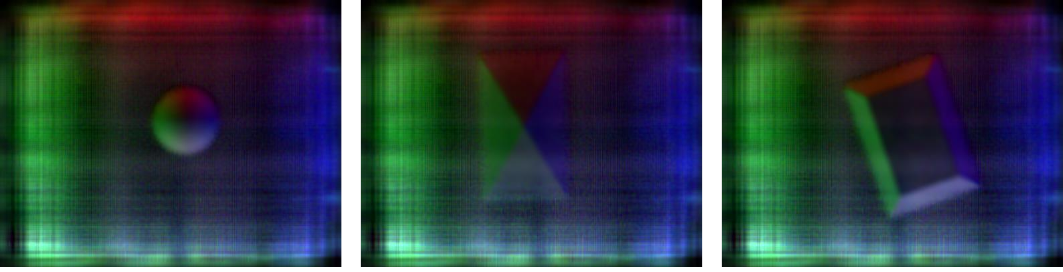}
		\label{fig:sim_indentation}
	}
     \subfigure[]{
		\includegraphics[width=7cm]{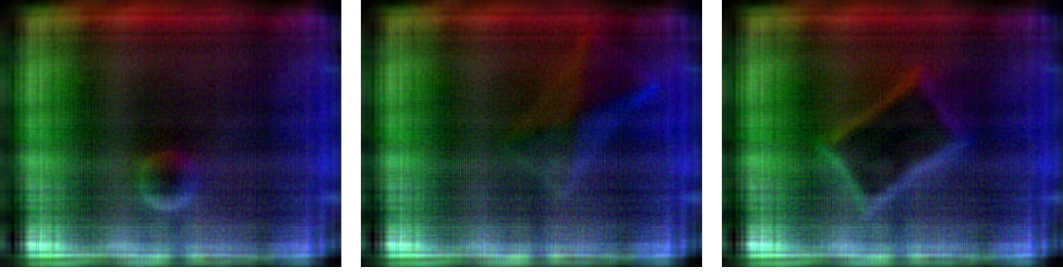}
		\label{fig:real_indentation}
	}
      \subfigure[]{
		\includegraphics[width=7cm]{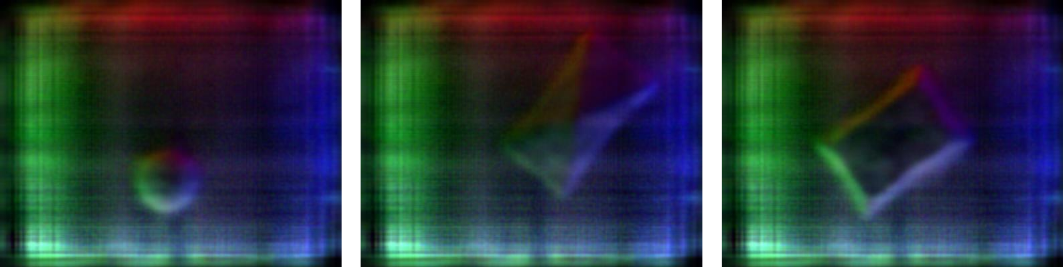}
		\label{fig:real2sim_indentation}
	}
    \caption{Indentation experiment for evaluating the depth reconstruction accuracy of \name. (a) Indenters used for the experiment. 8 indenters with different shapes are used while 6 of them are used for training and the remaining are used for test. The real indenters are 3D printed using stereo lithography. (b) Setup for the indentation experiment. (c) Simulated tactile images. (d) Real tactile images. (e) Real2Sim tactile images which are generated using the trained Real2Sim generator from CycleGAN.}
    \label{fig:indentation_experiment}
\end{figure}

As mentioned in Sec. \ref{sec:depth_recon}, we employ a Real2Sim transfer method to mitigate the non-ideal factors present in reality and improve the depth reconstruction quality. 

First, we follow~\cite{chen2022bidirectional} to collect the unpaired synthetic and real data. We design 8 indenters with various shapes, as depicted in Fig. \ref{fig:indenters}. For the real dataset, the tactile sensor remains fixed, while an indenter is secured on a 4-DoF stage (3 translational DoF and 1 rotational DoF), as shown in Fig. \ref{fig:indentation_setup}. For each indenter, we randomize its position and rotation, and set its indentation depths to 0.25 mm, 0.50 mm, 0.75 mm, and 1.00 mm, respectively. We collect 50×4=200 images per indenter.
For the simulation, we generate the surface depth map by intersecting the indenter mesh with the surface mesh. The depth map is then processed using Gaussian filters to approximate the deformation of the elastomer~\cite{gomes2021generation}. Subsequently, we employ Phong’s Shading to generate ideal tactile images that perfectly meet the requirements of photometric stereo. These simulated images are then blended with the background of the real images. For each indenter, we also generate 50×4=200 images with 4 indentation depths, with randomized positions that are not aligned with the real dataset. Several real and simulated images are displayed in Fig. \ref{fig:indentation_experiment}.

We use 6 indenters for training CycleGAN and the remaining 2 indenters for test. Calibration of the LUT is performed in advance using a 3 mm sphere indenter, and the calibration procedure is performed separately using real images and the Real2Sim images. We present several Real2Sim tactile images in Fig. \ref{fig:real2sim_indentation}. As can be seen in the figure, real tactile images suffer from a noisy background and non-uniform illumination. In contrast, the Real2Sim images exhibit more idealness than the real ones. We further evaluate our sensor's depth reconstruction accuracy quantitatively. To calculate the depth error, we generate the ground-truth depth using simulation and apply thresholding to select the area for error calculation. The depth error is calculated by first calculating the root-mean-squared error of all the pixels in the aforementioned area in each image, and then averaging the errors across all the tactile images. The results are summarized in Table \ref{tab:depth_accuracy}. The data shows that direct reconstruction from real tactile images results in a large depth error due to non-ideal conditions. In contrast, the Real2Sim method achieves an error of 0.13 mm and improves the depth accuracy by approximately 60\%.

\begin{table}[htbp]
  \centering
  \caption{Comparison of Depth Reconstruction Error From Real Tactile Images and From the Real2Sim Ones}
    \begin{tabular}{cccc}
    \toprule
    Indenter name & Indentation depth & Real  & Real2Sim \\
    \midrule
    \multirow{4}[8]{*}{chamfered prism} & 0.25 mm & 0.248 mm& \textbf{0.065} mm\\
\cmidrule{2-4}          & 0.50 mm  & 0.341 mm& \textbf{0.090} mm \\
\cmidrule{2-4}          & 0.75 mm & 0.407 mm& \textbf{0.098} mm\\
\cmidrule{2-4}          & 1.00 mm & 0.447 mm& \textbf{0.142} mm\\
    \midrule
    \multirow{4}[8]{*}{4 mm sphere} & 0.25 mm & 0.262 mm& \textbf{0.169} mm\\
\cmidrule{2-4}          & 0.50 mm & 0.324 mm & \textbf{0.172} mm\\
\cmidrule{2-4}          & 0.75 mm & 0.365 mm& \textbf{0.156} mm \\
\cmidrule{2-4}          & 1.00 mm & 0.365 mm& \textbf{0.109} mm\\
    \bottomrule
    \end{tabular}%
  \label{tab:depth_accuracy}%
\end{table}%

\subsection{Fabric Texture Classification}\label{sec:texture_recognition}
To evaluate the capability of \name\ to capture fine details, we conduct an experiment of fabric texture classification, where we use a robot arm to press the tactile sensor against various fabric materials. A deep neural network is trained to classify the reconstructed tactile images.

\begin{figure*}
    \includegraphics[width=\linewidth]{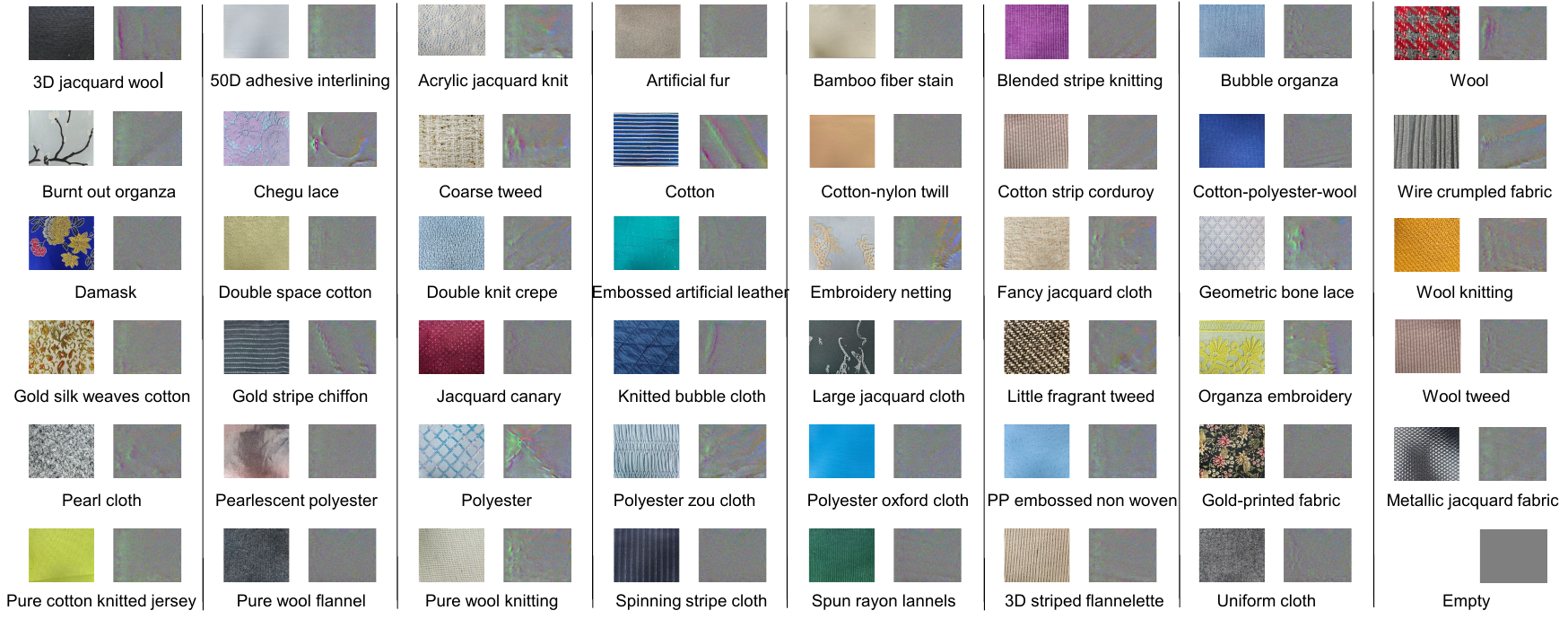}

    \caption{RGB photos and reconstructed tactile images of the 47 kinds of fabrics used for fabric texture classification.}
        \label{fig:texture}
\end{figure*}

\noindent \textbf{Fabric materials.} 
We select 47 kinds of fabrics for fabric texture classification. These fabrics are composed of diverse materials such as cotton, linen, silk, polyester, spandex, or wool, resulting in nuances of tactile signals. As shown in Fig.~\ref{fig:texture}, \name\ is able to reconstruct the subtle fabric texture geometries.

\begin{figure}
    \centering
     \subfigure[]{
		\includegraphics[width=4.8cm]{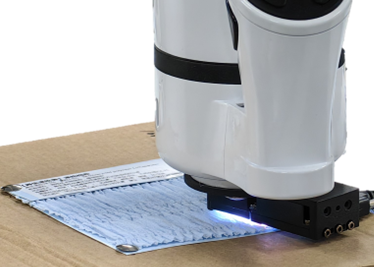}
		\label{fig:texture_arm}
	}
    \hfill
    \subfigure[]{
		\includegraphics[width=5.8cm]{figures/confusion_matrix.pdf}
		\label{fig:confusion_matrix}
	}
    
    \caption{Fabric texture classification. (a) Dataset collection setup. (b) Confusion matrix. It is shown that most materials have a 100\% classification accuracy, indicating that \name\ can accurately identify subtle texture differences.}
    \label{fig:texture_classification}
\end{figure}

\noindent \textbf{Data Collection and Network.} 
To construct the dataset, we collect 100 images for each texture and divide them into three subsets: a training set, a validation set, and a testing set, with a ratio of 6:2:2. 
As shown in Fig.~\ref{fig:texture_arm}, we affix \name~to the end of a ROKAE xMate3-pro robotic arm. Starting from a corner, the arm is programmed to move along horizontal and vertical directions with a step size of 8 mm and rotate 5° on the surface of each fabric sample. The size of each fabric sample is 10 cm $\times$ 10 cm. The applied pressure remains approximately constant across all fabric samples. After each contact, a measurement image was captured and reconstructed to produce a tactile image with a resolution of $512 \times 400$. 

We use ResNet-50 image classification network~\cite{he2016deep}, and train the network for 100 epochs with a batch size of 64 and a learning rate of 1e-3.

\noindent \textbf{Results and data analysis.}
The classification precision on the test set is 98.96\%. Fig.~\ref{fig:confusion_matrix} shows the confusion matrix, where most fabrics have a 100\% classification accuracy, demonstrating the sensor's ability to recognize subtle surface texture features.

\section{Application in Manipulation}\label{sec:applicatoin}
Tactile sensing offers significant benefits for robots, particularly in terms of the ability to interact with unstructured environments. In these settings, objects and devices are typically designed for human use, and thus, are well-suited to the dimensions of human fingertips, which generally range from 15 to 20 millimeters in diameter. Consequently, tactile sensors with larger dimensions may struggle to accommodate such constrained spaces. In this section, we first present \name's sensitivity when handling delicate objects, and then present several instances where tactile sensing is essential and space is limited, to prove the advantages of \name. Demonstrations are available in the supplementary video associated with this article.

\subsection{Grasping of Delicate Objects}

\begin{figure}
    \centering
    \subfigure[]{
		\includegraphics[width=4.1cm]{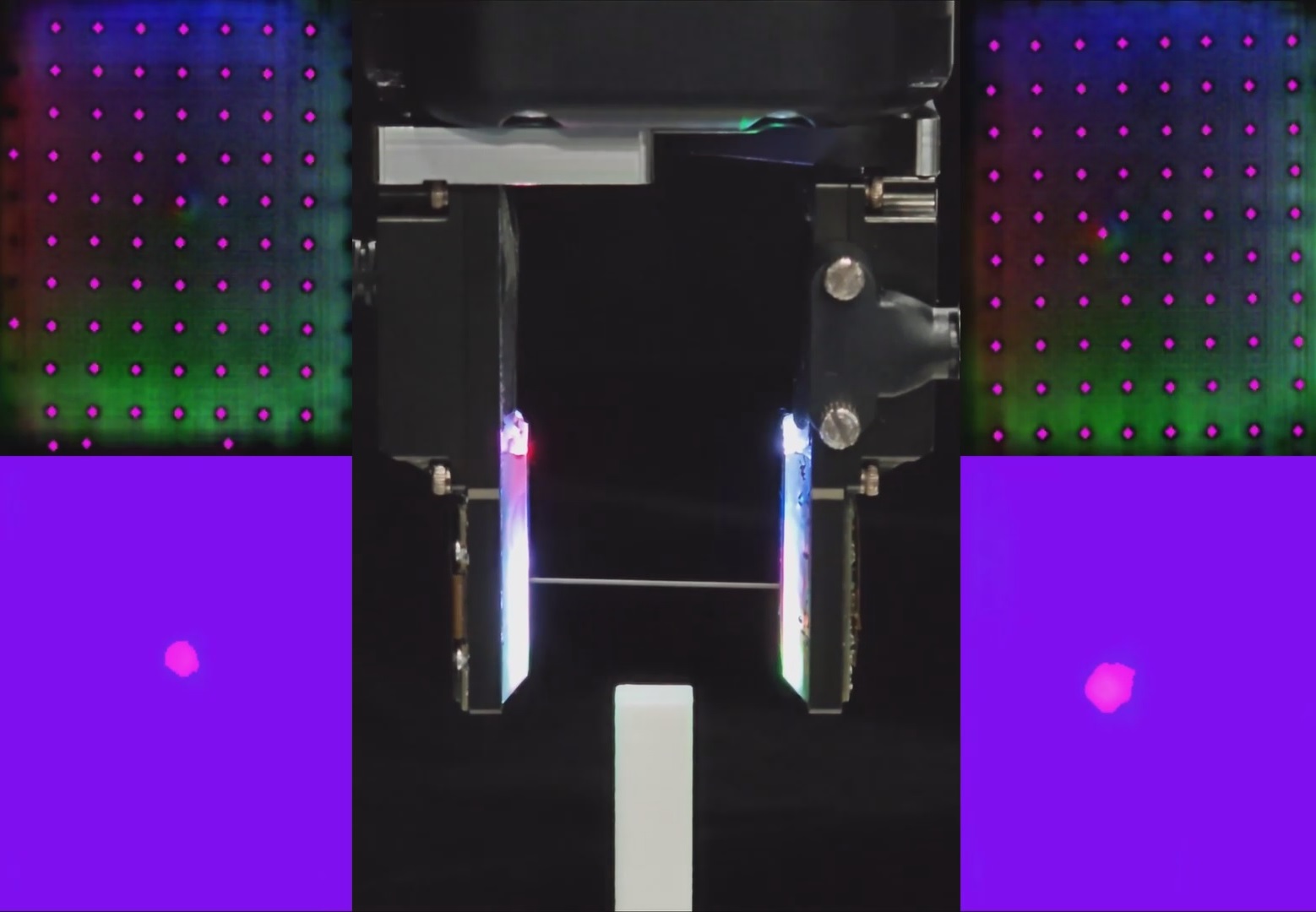}
		\label{fig:grasp_pencil_lead}
	}
    \hfill
    \subfigure[]{
		\includegraphics[width=4.1cm]{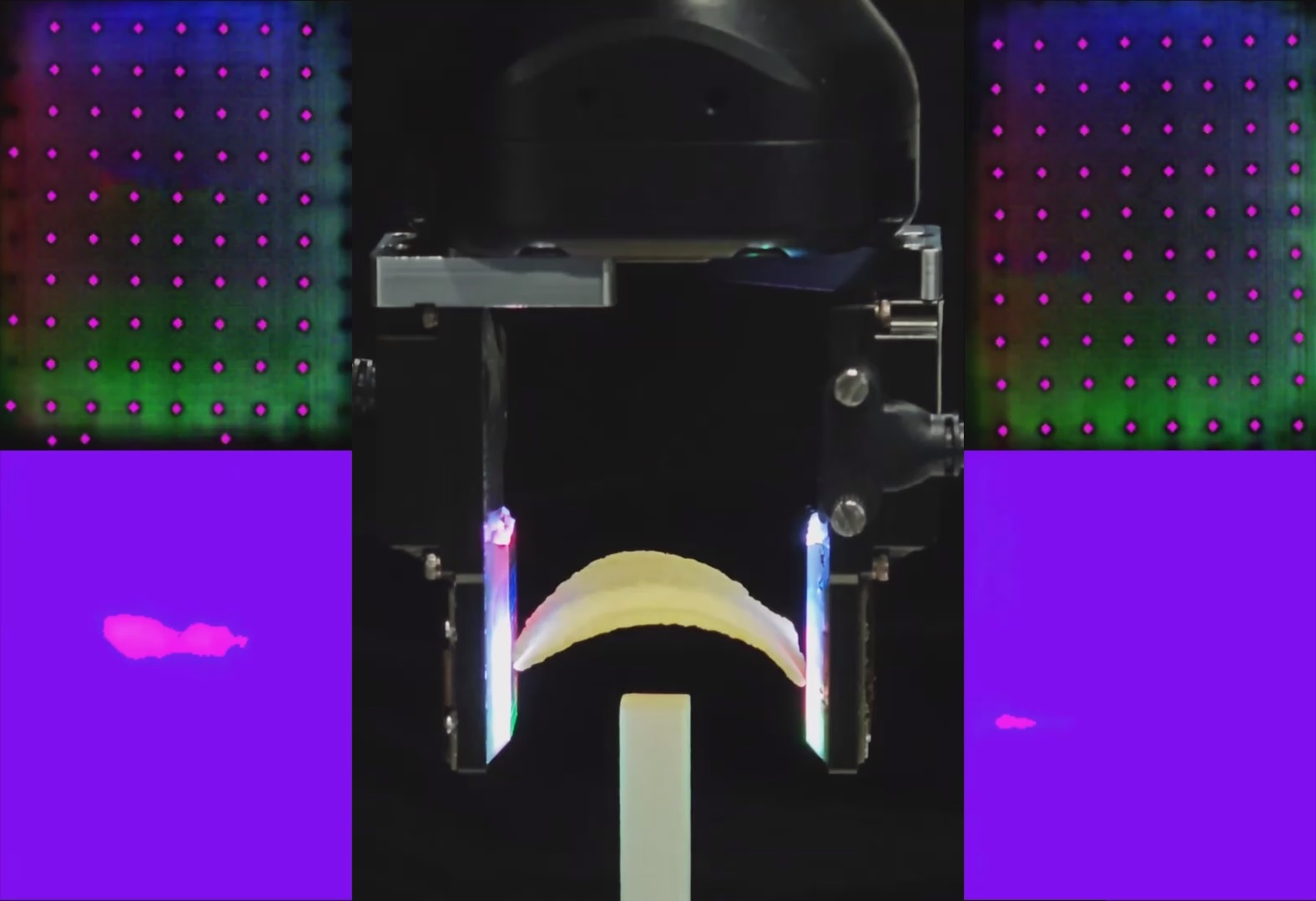}
		\label{fig:grasp_potato_chip}
	}    
    \caption{\name~ grasps two delicate objects. On both sides of each image are the tactile readings of the sensors, and the upper ones are the marker flow image while the lower ones are the depth images. In the depth images, the area where the depth is larger than 0.1 mm is highlighted in red.}
    \label{fig:grasp_delicate_objects}
\end{figure}

To demonstrate the sensitivity of \name, we conduct experiments to grasp two delicate objects: a 0.5 mm pencil lead and a potato chip. In the experiment, the fingertips of the parallel gripper gradually closes, and the gripper will immediately stop once a tactile sensor detects contact. Fig. \ref{fig:grasp_delicate_objects} shows that these two delicate objects are successfully grasped and lifted by \name. The corresponding tactile images are also shown in Fig. \ref{fig:grasp_delicate_objects}. We use a high-precision force sensor (Rtec MFT-5000) with a resolution of 1.5 mN to measure the force applied when the sensor detects contact. The result shows that the force is 0.16 N when a contact is detected, demonstrating the high sensitivity of our proposed tactile sensor.

\begin{figure}
    \centering
	\includegraphics[width=8.5cm]{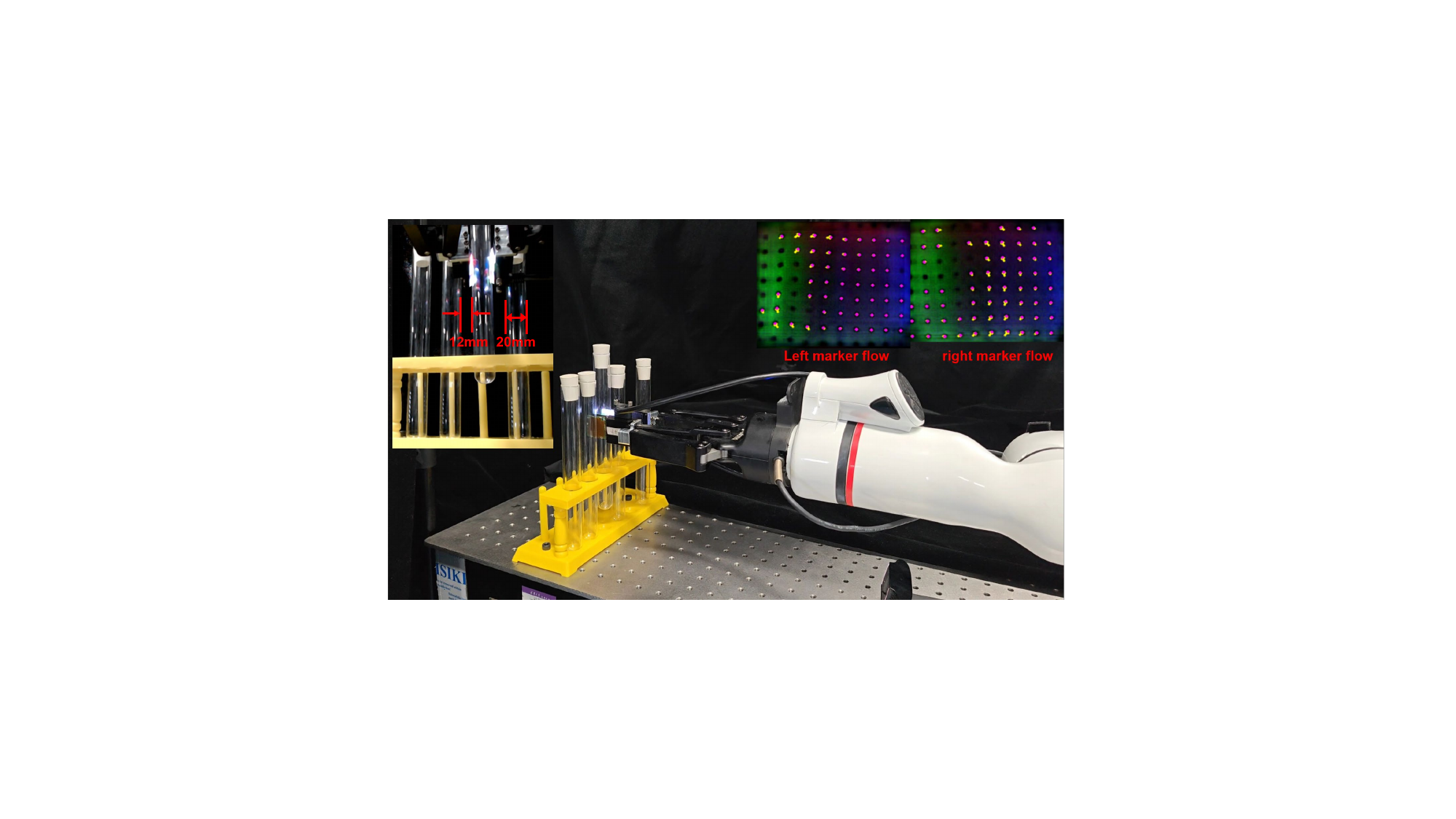}
    \caption{\name\ extracts tactile marker flow signals to adjust the positioning of test tubes in real time, allowing them to be placed into a test tube rack. The diameter of the tube is 20 mm. The distance between two tubes is only 12 mm and the clearance between the tube and the hole is 1 mm. The marker flow images are shown in the upper right corner.}
    \label{fig:insert_tube}
\end{figure}

\subsection{Insertion of Test Tubes}

Using \name\ for perception and feedback control, we are able to accomplish tasks such as precisely inserting test tubes into a test tube rack, where the diameter of the tube is 20 mm, the diameter of the hole is 22 mm, the distance between two tubes is only 12 mm, and the clearance of the hole is 1 mm. As shown in Fig.\ref{fig:insert_tube}, we utilize a Robotiq 2F-140 gripper fixed at the end of a ROKAE xMate3-pro robotic arm for grasping. Two \name s are installed on both sides of the gripper. 

To simulate the positioning errors from vision perception or human handover, we introduce a random offset to the initial pose of the test tube. When the test tube makes contact with the rack during the placement, we capture the marker displacements from the sensors and fit them to derive the direction of the force exerted on the test tube. Subsequently, we adjust the gripper's pose in the opposite direction, allowing the test tube to be precisely inserted into the hole. For a detailed demonstration, please refer to the accompanying video.

\subsection{Daily Object Manipulation}
\begin{figure}
    \centering
    \subfigure[]{
		\includegraphics[height=2.0cm]{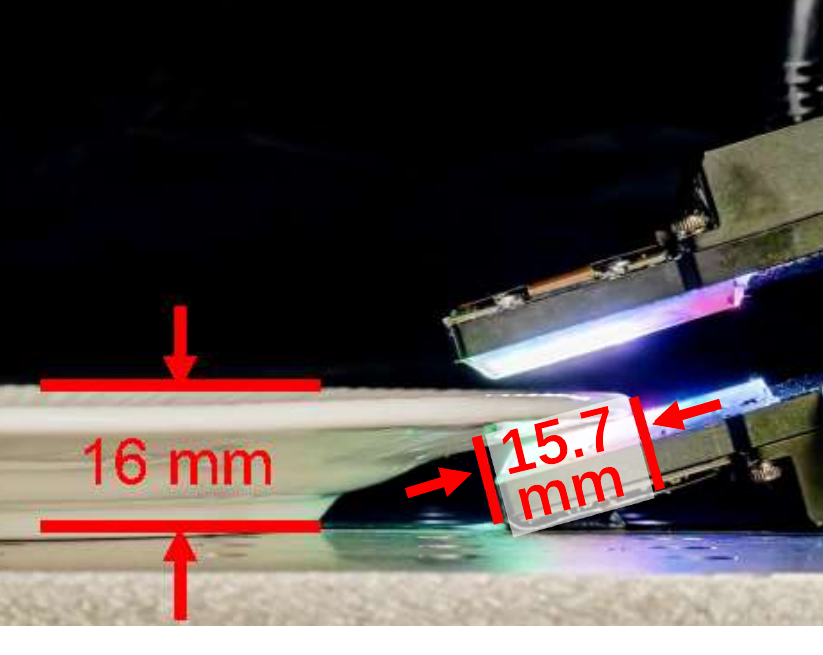}
		\label{fig:tinsel_with_plate}
	}
    \hfill
    \subfigure[]{
		\includegraphics[height=2.0cm]{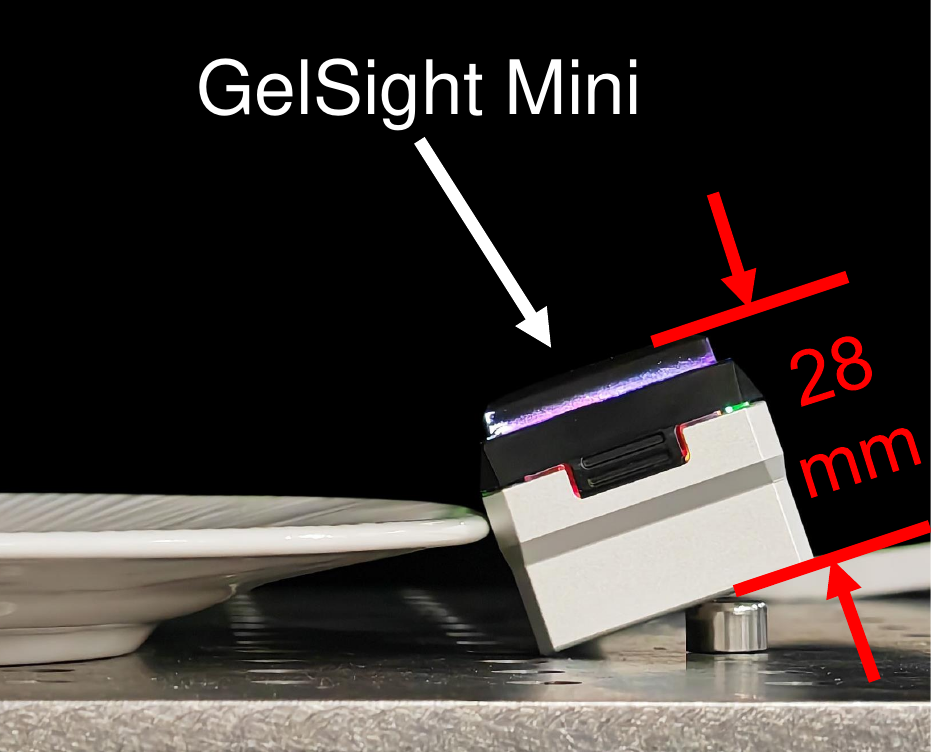}
		\label{fig:gelsight_mini_with_plate}
	}
    \hfill
    \subfigure[]{
    \includegraphics[height=2.0cm]{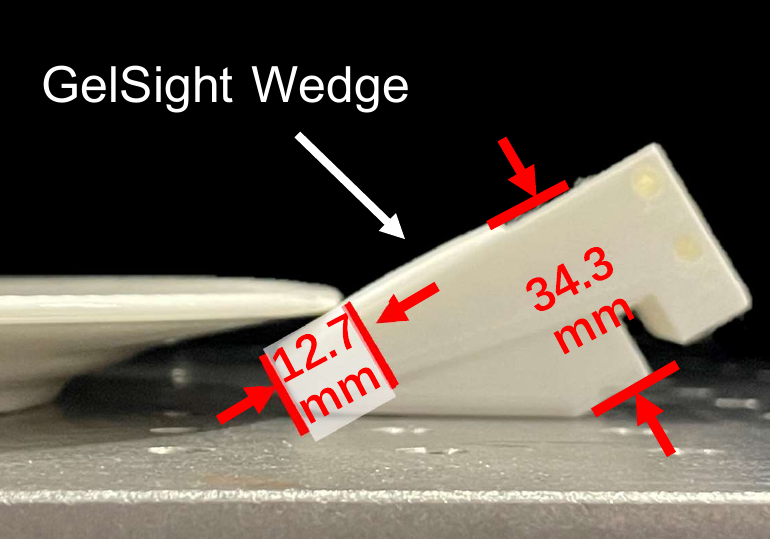}
		\label{fig:gelsight_wedge_with_plate}
    }
    \caption{\name~grasps a flat plate. (a) \name~is able to fit into the gap at the boundary. (b) GeiSight Mini is too thick to grasp the plate. (c) GelSight Wedge is also able to fit into the gap, but the insertion depth is smaller. The video for this task is available in the multimedia attachment.}
    \label{fig:grasp_plate}
\end{figure}

\noindent \textbf{Grasp a plate.}
In this experiment, the object of interest is a flat plate with a height of only 16 mm. To grasp such a thin plate, a slim fingertip is required. As depicted in Fig. \ref{fig:grasp_plate}, \name~successfully inserts itself into the gap at the edge of the plate. In contrast, GeiSight Mini fails to do so due to its thickness exceeding the height of the plate. While GelSight Wedge~\cite{wang2021gelsight} can also be inserted into the gap, its insertion depth is limited compared to ours. This limitation is due to the mirror angle, which cannot be further reduced to provide the appropriate reflection into the camera at the bottom. Tactile sensing is employed to determine the plate’s position and trigger the gripping action. The gripper, Robotiq Hand-E, initially moves along the tabletop towards the plate until \name~detects contact with the plate. Subsequently, the gripper closes, enabling the robot to lift the plate.

\begin{figure}
    \centering
    \subfigure[]{
		\includegraphics[height=4cm]{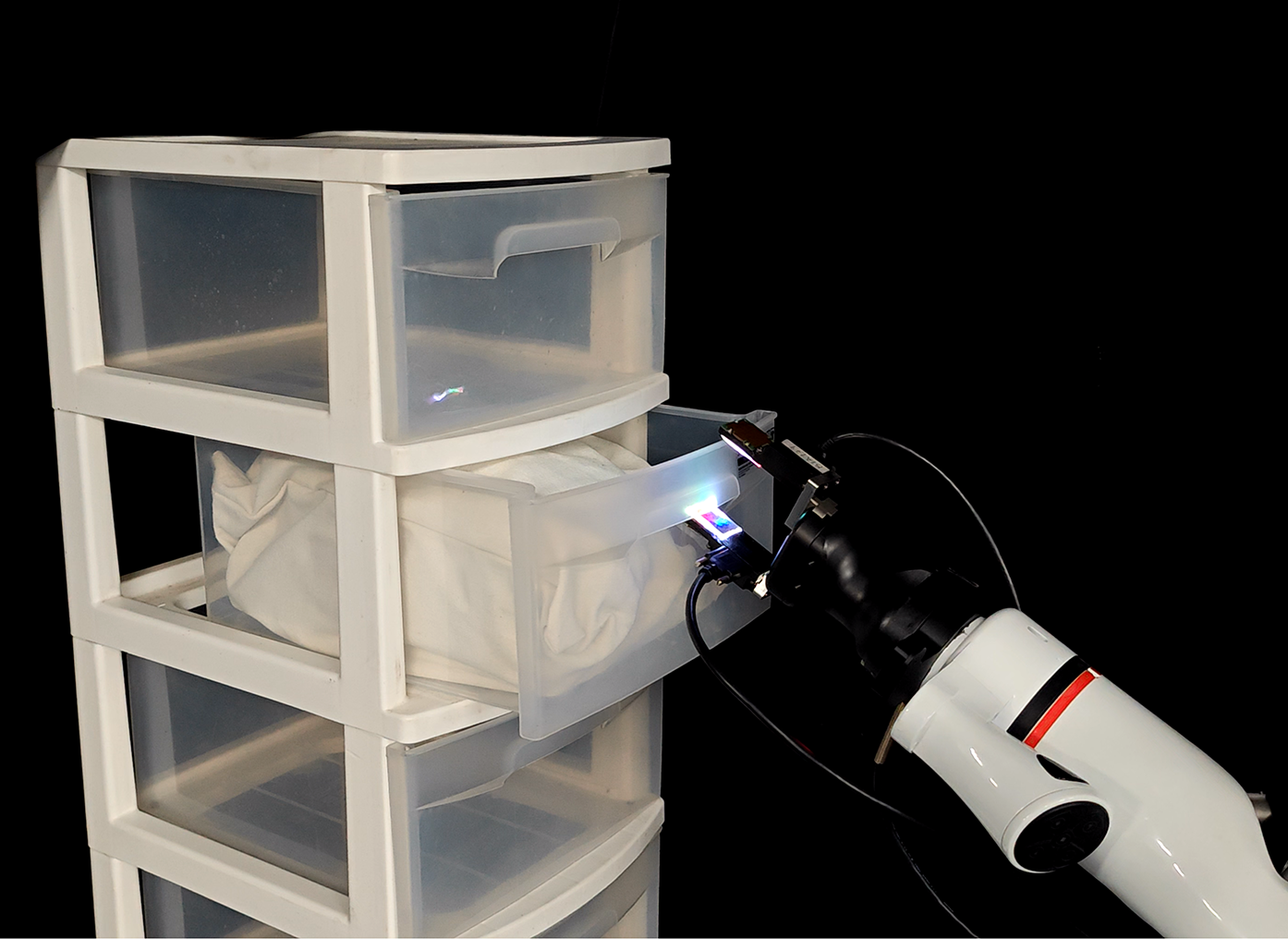}
		\label{fig:tinsel_open_drawer}
	}
    \hfill
    \subfigure[]{
		\includegraphics[height=4cm]{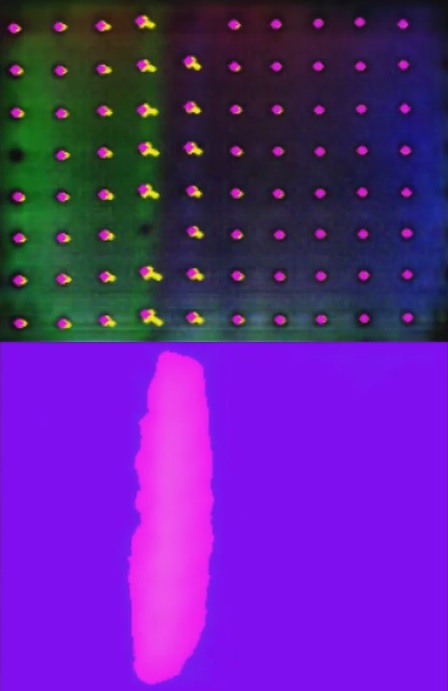}
		\label{fig:open_drawer_tactile_reading}
	}
 
    \subfigure[]{
		\includegraphics[height=2.9cm]{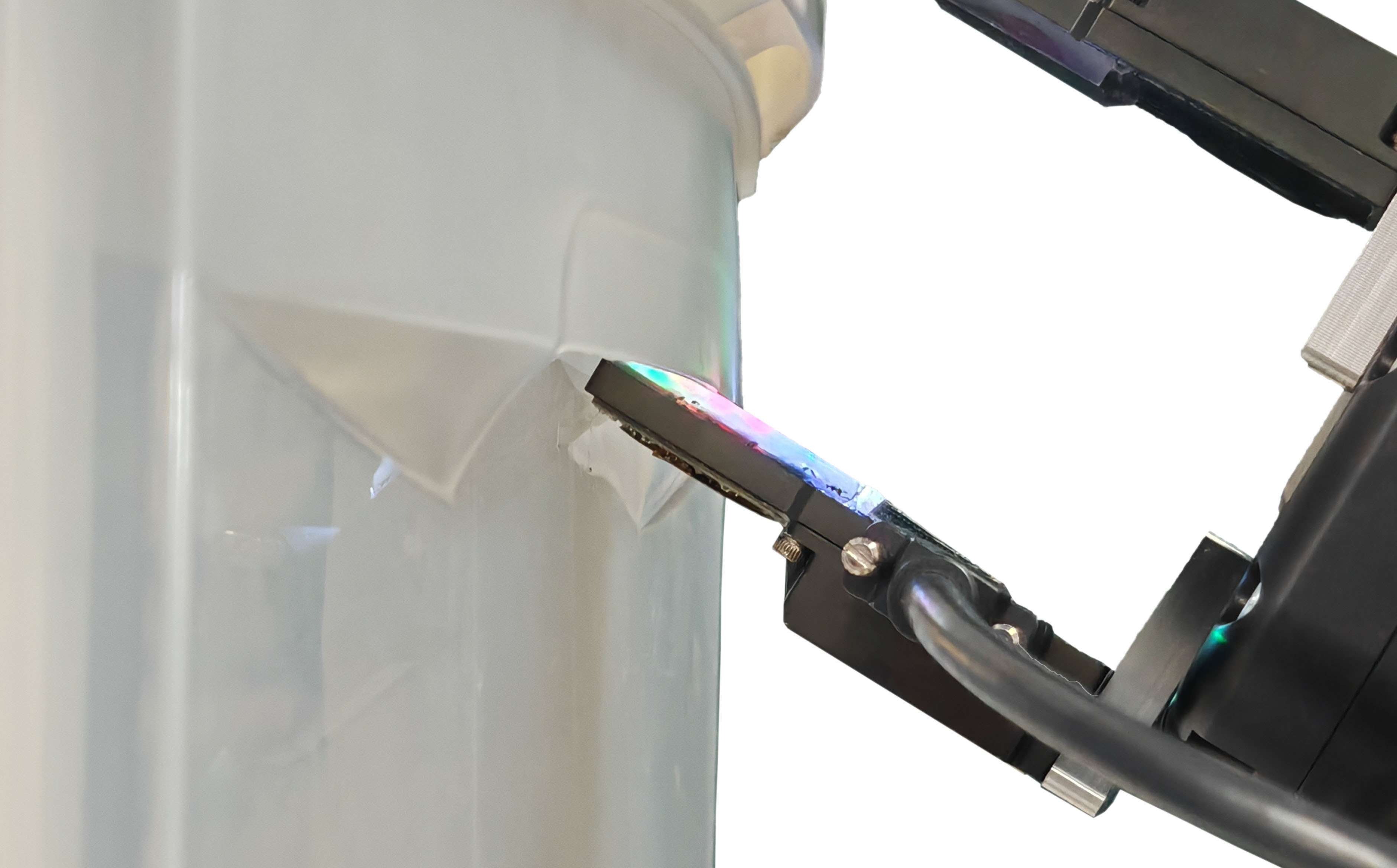}
		\label{fig:tinsel_with_drawer}
	}
    \hfill
    \subfigure[]{
		\includegraphics[height=2.9cm]{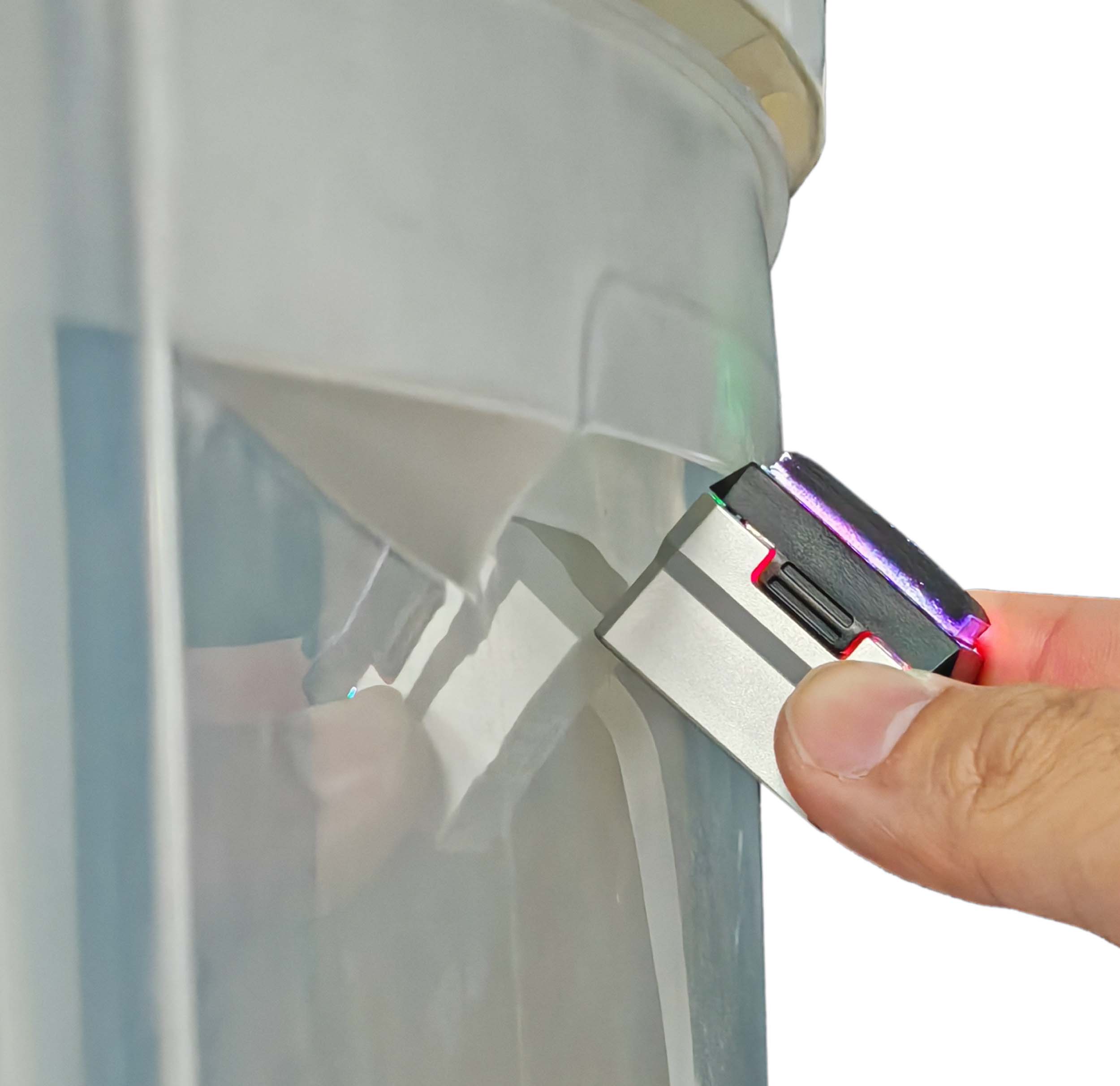}
		\label{fig:gelsight_mini_with_drawer}
	}

    \caption{Drawer opening experiment. (a) \name~successfully opens the drawer. (b) Tactile readings of \name~when opening the drawer. (c) \name~is able to fit into the handle and detect the contact. (d) GelSight Mini is too thick to fit into the handle. The video for this task is available in the multimedia attachment.}
    \label{fig:open_drawer}
\end{figure}

\noindent \textbf{Open the drawer.}
In this task, the robot is required to locate the drawer’s handle and exert the necessary force to open it. The handle is designed with a downward opening, and the width of this opening is less than 20 mm, suitable for human use. To successfully open the drawer, the direction of contact should be outward to apply the correct forces, as illustrated in Fig. \ref{fig:tinsel_with_drawer}. Given these constraints, tactile sensors with a larger thickness, such as GelSight Mini, are unable to accomplish the task (Fig. \ref{fig:gelsight_mini_with_drawer}). In contrast, \name~first detects contact with the handle through an upward movement, fits into the handle, and ultimately completes the task (Fig. \ref{fig:tinsel_open_drawer}).

\section{Discussion and Conclusion}\label{sec:sec_conclusion}

In this work, we present \name, a thin, vision-based tactile sensor with a substantial sensing area, powered by lensless imaging. Our extensive experiments demonstrate its ability to reconstruct contact geometries, recognize textures, and manipulate a variety of everyday objects. It performs comparably to traditional GelSight-type sensors, but with a significantly reduced thickness, making it suitable for integration into robots and use in confined spaces.

Despite the promising results, there are several limitations and directions for future investigation. First, the sensor is not thin enough to fully reveal the advantage of lensless imaging. This may be attributed to that the CMOS we use is over 4 mm thick itself, while current technology is already capable of fabricating large-area CMOS with thickness less than 1 mm, such as the ones used in smartphones. Second, although the part of the sensor beneath the sensing surface is thin, the sensor also has a thicker body part, which makes it currently unsuitable for integration into a dexterous hand. This issue could be addressed by developing customized, small-sized image processing units, or by considering all sensors in a dexterous hand and placing the multi-input image processing unit in an appropriate location. Finally, like other camera-based tactile sensors, \name~suffers from a limited sampling rate. To address this, one could consider incorporating other tactile sensing modalities, using high-end, fast CMOS sensors, or exploring event-based image sensors.

% \balance
\bibliographystyle{IEEEtran}
\bibliography{references}

\begin{IEEEbiography}[{\includegraphics[width=1in,height=1.25in,clip,keepaspectratio]{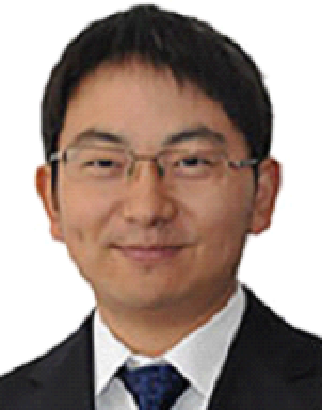}}]{Jing Xu}
received his Ph.D. in mechanical engineering from Tsinghua University, Beijing, China, in 2008. He was a Postdoctoral Researcher in the Department of Electrical and Computer Engineering, Michigan State University, East Lansing. He is currently an Associate Professor in the Department of Mechanical Engineering, Tsinghua university, Beijing, China. His research interests include vision-guided manufacturing, image processing, and intelligent robotics.
\end{IEEEbiography}

\begin{IEEEbiography}[{\includegraphics[width=1in,keepaspectratio]{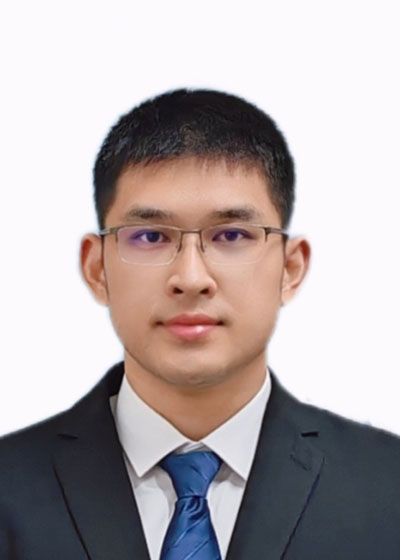}}]{Weihang Chen} received his B.E. degree in mechanical engineering in 2018 from Tsinghua University, Beijing, China, where he is working toward his Ph.D. in the Department of Mechanical Engineering. His research interests include vision-based tactile sensing and Sim2Real for robotics.  
\end{IEEEbiography}

\begin{IEEEbiography}[{\includegraphics[width=1in,height=1.25in,clip,keepaspectratio]{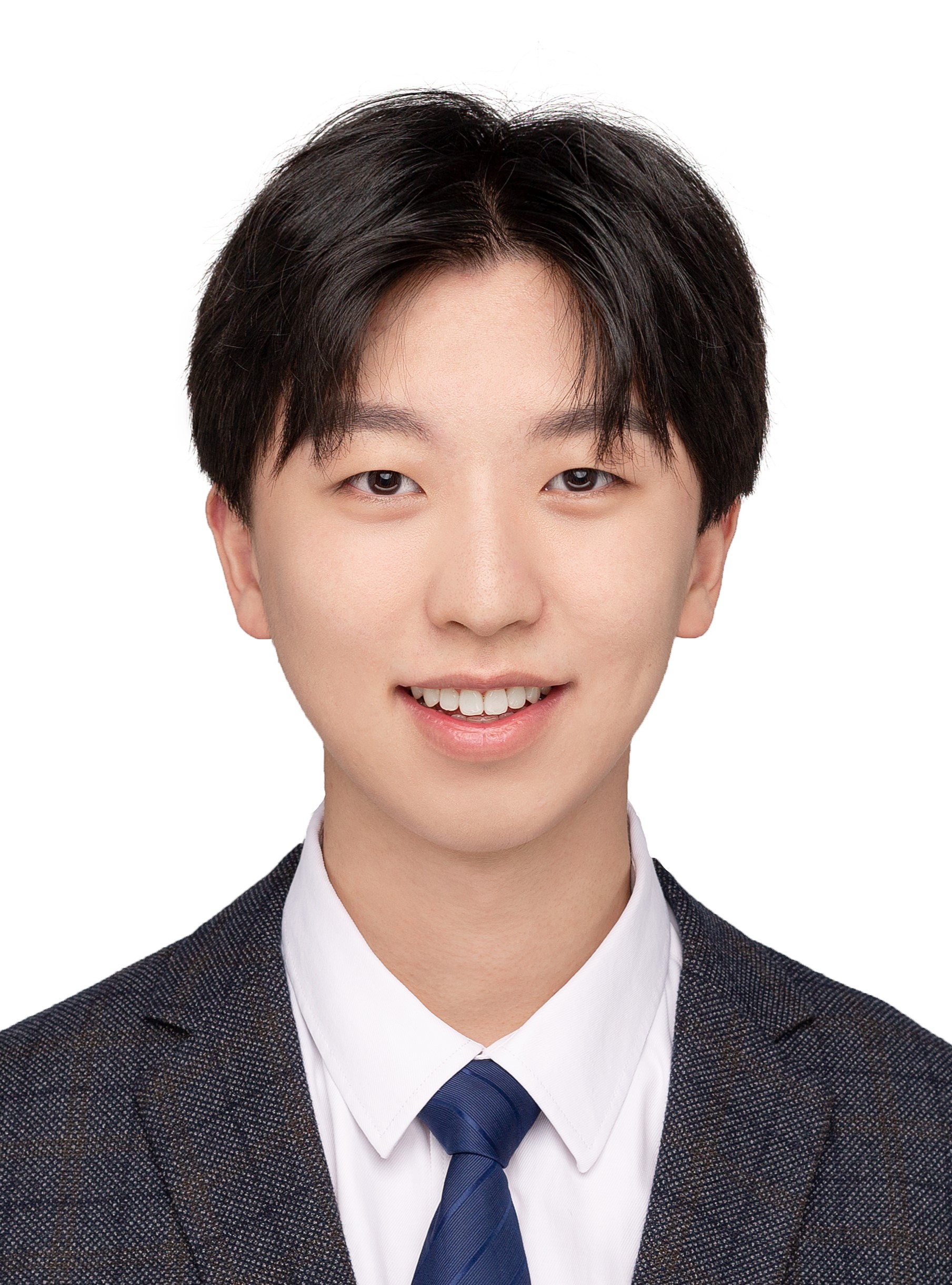}}]{Hongyu Qian}
is currently a Ph.D. student in the Department of Mechanical Engineering at Tsinghua University. He obtained his B.E. degree from Southwest Jiaotong University in 2023. His current research focuses on robotics and tactile sensors.
\end{IEEEbiography}

\begin{IEEEbiography}[{\includegraphics[width=1in,height=1.25in,clip,keepaspectratio]{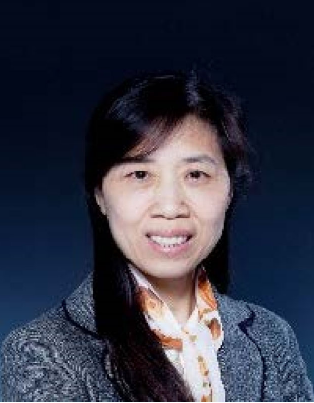}}]{Dan Wu}
received the B.S., M.S., and Ph.D. degrees in mechanical engineering from Tsinghua University of Beijing, China in 1988, 1990, and 2008, respectively. Since 1990, she has been working in Tsinghua University. Now she is a Full Professor with the Department of Mechanical Engineering,
Tsinghua University. She has authored or coauthored
more than 160 papers. Her research interests include robotic manipulation, precision motion control, precision and ultra-precision machining. Prof. Wu is a member of American Society for Precision Engineering and a senior member of the Chinese Mechanical Engineering Society.
Also, she is an associate editor of the Journal of Intelligent Manufacturing and Special Equipment.
\end{IEEEbiography}

\begin{IEEEbiography}[{\includegraphics[width=1in,height=1.25in,clip,keepaspectratio]{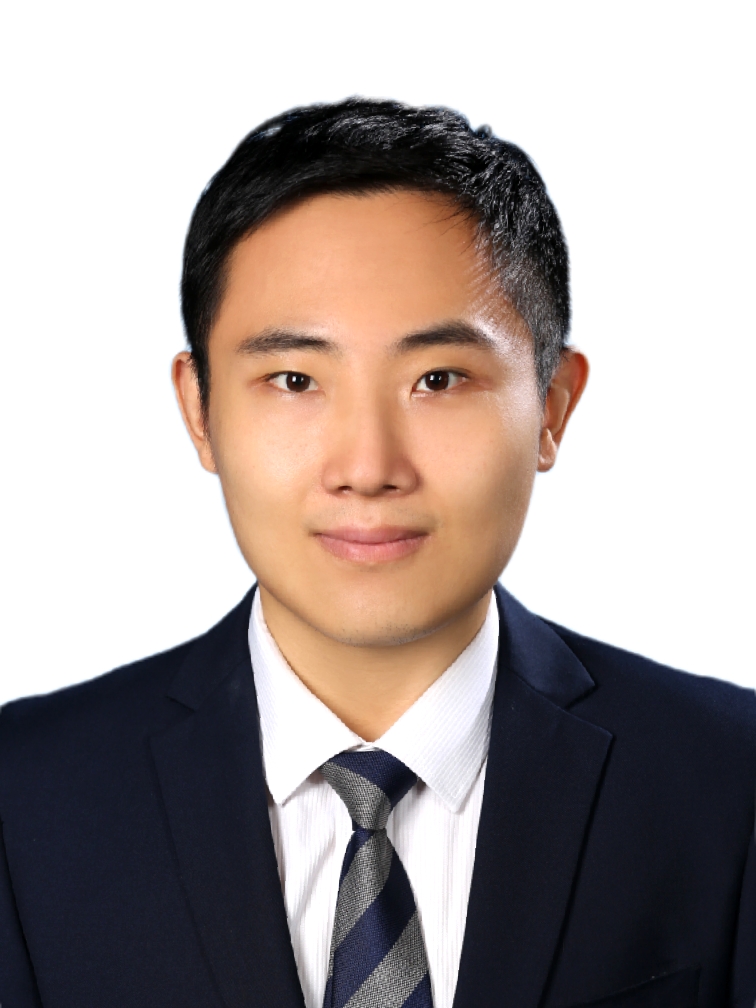}}]{Rui Chen} is currently a research assistant professor in the Department of Mechanical Engineering, Tsinghua University. He received the Ph.D. degree in mechatronical engineering and the B.E. degree in mechanical engineering in 2020, 2014 from Tsinghua University, Beijing, China. His research interests include tactile sensing, three-dimensional computer vision, and robot learning.
\end{IEEEbiography}
\vfill
\end{document}